\documentclass[10pt]{article}

\usepackage[preprint]{tmlr}

\usepackage{hyperref}
\usepackage{url}

\usepackage{algorithm}
\usepackage{algpseudocode}

\usepackage{graphicx}
\usepackage{booktabs,enumitem} 
\usepackage{makecell}
\usepackage{multirow}
\usepackage{caption}

\usepackage{amsfonts}       
\usepackage{amsmath}
\usepackage{amssymb}
\usepackage{mathtools}
\usepackage{amsthm}
\usepackage{bm}

\theoremstyle{plain}
\newtheorem{theorem}{Theorem}

\newtheorem{lemma}{Lemma}
\newtheorem{corollary}{Corollary}
\theoremstyle{definition}

\newtheorem{assumption}{Assumption}
\theoremstyle{remark}
\newtheorem{remark}{Remark}

\newcommand{\op}[1]{\operatorname{#1}}
\newcommand{\C}[1]{\mathcal{#1}}
\newcommand{\BF}[1]{\mathbf{#1}}
\newcommand{\BB}[1]{\mathbb{#1}}
\newcommand{\K}{\C{K}}

\newcommand{\x}{\BF{x}}
\newcommand{\xt}{\x_t}
\newcommand{\xs}{\x^*}
\newcommand{\txs}{\tilde{\x}^*}
\newcommand{\y}{\BF{y}}

\newcommand{\z}{\BF{z}}

\newcommand{\g}{\BF{g}}
\newcommand{\tf}{\tilde{f}}
\newcommand{\tft}{\tilde{f}_t}
\newcommand{\tg}{\tilde{g}}
\newcommand{\tgt}{\tilde{g}_t}
\newcommand{\hf}{\hat{f}}
\newcommand{\hft}{\hat{f}_t}
\newcommand{\normdt}{\Vert \nabla_t \Vert}
\newcommand{\normdm}{\Vert \bn_m \Vert}
\newcommand{\sumT}{\sum\limits_{t=1}^T}
\newcommand{\qt}{Q_t}
\newcommand{\qtt}{Q_{t-1}}
\newcommand{\bn}{\bar{\nabla}}
\renewcommand{\cite}[1]{\citep{#1}}

\title{BAGEL: Adversarially Constrained Online Convex \\ Optimization under Separation Oracle Access}

\author{\name Yiyang Lu \email yiyanglu@purdue.edu \\
  \addr Purdue University, West Lafayette, IN, USA
  \AND
  \name Mohammad Pedramfar \email mohammad.pedramfar@mila.quebec \\
  \addr Mila - Quebec AI Institute/McGill University, Montreal, QC, Canada
  \AND
  \name Mengbo Wang \email wang4887@purdue.edu \\
  \addr Purdue University, West Lafayette, IN, USA
  \AND
  \name Vaneet Aggarwal \email vaneet@purdue.edu \\
  \addr Purdue University, West Lafayette, IN, USA
}

\begin{document}

\maketitle

\begin{abstract}
In adversarial Constrained Online Convex Optimization (COCO), a learner selects actions from a fixed convex set while seeking both low regret and low cumulative constraint violation (CCV) under time-varying constraints. We ask what performance is achievable when the action set is accessed through a Separation Oracle (SO), rather than an exact Projection Oracle (PO) or a Linear Optimization Oracle (LOO). We introduce $\mathtt{BAGEL}$, which combines a Lyapunov-weighted surrogate loss, blocked adaptive online gradient descent, and an infeasible-projection procedure implemented with an SO. For convex costs and any $\beta\in(0,1/2]$, $\mathtt{BAGEL}$ achieves $\mathcal{O}(T^{1-\beta})$ regret and $\mathcal{O}(T^{1-\beta}\log T)$ cumulative violation using $\widetilde{\mathcal{O}}((D/r)^2T^{2\beta})$ SO calls. At $\beta=1/2$, this gives $\mathcal{O}(\sqrt{T})$ regret and $\mathcal{O}(\sqrt{T}\log T)$ violation with a near-linear number of SO calls. The result is an access oracle based guarantee, with computational relevance depends on the geometry of the action set and the cost of implementing its SO.
\end{abstract}

\section{Introduction}

Online Convex Optimization (OCO) has emerged as a foundational framework for sequential decision-making under uncertainty, with applications spanning resource allocation, real-time control systems, and adversarial robustness in machine learning \cite{shalev2012online}. A critical variant of this framework, Constrained Online Convex Optimization (COCO), requires agents to minimize cumulative costs while controlling violations of time-varying constraints. It models applications spanning diverse domains, including portfolio optimization under dynamic risk constraints \cite{redeker2018portfolio}, dynamic resource allocation for time-varying workloads \cite{doostmohammadian2022distributed}, real-time pricing with inventory constraints \cite{li2024dynamic}, and collision-free trajectory planning with safety constraints \cite{da2019collision}.

\paragraph{Problem Setup and Metrics.} We consider adversarial COCO over a horizon of $T$ rounds. Let $\K\subseteq\BB{R}^d$ be a fixed convex action set. At each round $t\in[T]$, the learner selects an action $\x_t\in\K$, before the adversary reveals a convex cost function $f_t:\K\to\BB{R}$ and $k$ convex constraint functions $g_{t,i}:\K\to\BB{R}$, $i\in[k]$, together with their first-order information at the played action. Let $\K^\star \triangleq \{ \x\in\K: g_{t,i}(\x)\leq 0 \forall t\in[T],\ i\in[k] \}$ denote the set of fixed actions that satisfy all constraints in hindsight. We assume $\K^\star\neq\varnothing$. The learner is evaluated using static regret and cumulative constraint violation:
\begin{align}
\mathtt{Regret}(T)
&=
\sum_{t=1}^T f_t(\x_t)
-
\min_{\x^\star\in\K^\star}
\sum_{t=1}^T f_t(\x^\star),
\\
\mathtt{CCV}(T)
&=
\max_{i\in[k]}
\sum_{t=1}^T
\bigl(g_{t,i}(\x_t)\bigr)^+,
\end{align}
where $(z)^+\triangleq\max\{z,0\}$. The goal is to achieve sublinear regret and CCV simultaneously.

\paragraph{Access Oracles.} The computational character of this problem depends on how the learner accesses the fixed action set $\K$. Projection-based methods use a Projection Oracle (PO), or a more general Convex Optimization Oracle (COO), whereas projection-free methods commonly use either a Linear Optimization Oracle (LOO) or a Separation Oracle (SO). Given a query point, an SO either certifies membership in $\K$ or returns a hyperplane separating the query point from $\K$. These oracle models are not uniformly ordered by computational cost. Their relative efficiency depends on the geometry and representation of $\K$. For instance, an LOO is highly efficient for a nuclear norm ball while a SO is considered efficient for the spectral norm ball because of rank-one SVD and full rank SVD, and they are both complementary natural ways to access the constrained action set \citep{garber2022new}. We therefore evaluate an algorithm not only by its regret and CCV, but also by the number of calls it makes to its action-set oracle. For a more detailed discussion we refer readers to the projection-free OCO works\citep{garber2022new, mhammedi2022efficient, mhammedi2025online}.

\paragraph{Research Landscape.} Existing adversarial COCO guarantees depend strongly on the access model. Under COO or PO access, algorithms achieve $\sqrt{T}$-scale regret while controlling cumulative violation \citep{guo2022online,yi2023distributed,sinha2024optimal}, and subsequent projection-based work has further refined the available regret--violation guarantees \citep{sinha2025beyond,vaze2025instance,sarkar2026geometric}. In contrast, existing projection-free methods based on an LOO typically obtain $T^{3/4}$-scale regret and violation guarantees \citep{garber2024projection,sarkar2025projection,wang2025revisiting}. Separation-oracle implementations of infeasible projection are known to achieve $\sqrt{T}$ regret in classical OCO \citep{garber2022new,mhammedi2025online}, but these results do not directly characterize the joint regret, constraint-violation, and query guarantees achievable in adversarial COCO.
This leads to the following oracle-theoretic question:
\begin{quote}
\emph{What performance is achievable in adversarial COCO when the action set is accessed through a separation oracle rather than an exact projection oracle or a linear optimization oracle?}
\end{quote}

Our answer is a regret-violation-oracle tradeoff. For convex costs and any $\beta\in(0,1/2]$, we give an algorithm that achieves $\mathtt{Regret}(T) =\C{O}(T^{1-\beta})$, $\mathtt{CCV}(T)=\C{O}(T^{1-\beta}\log T)$, and $\widetilde{\C{O}}\left((D/r)^2 T^{2\beta}\right)$ calls to an SO, where $D$ is the diameter of $\K$ and $r$ is an interior-radius parameter that captures the geometry required by the infeasible-projection procedure. At $\beta=1/2$, the algorithm obtains $\C{O}(\sqrt{T})$ regret and $\C{O}(\sqrt{T}\log T)$ CCV using a near-linear number of SO calls, up to geometry-dependent factors. Thus, under separation access, we achieve the $\sqrt{T}$ polynomial horizon dependence. This is an oracle-complexity statement rather than a universal claim that an SO has lower arithmetic or wall-clock cost than a PO or LOO.

\paragraph{Contributions.} We present $\mathtt{BAGEL}$ (\underline{B}locked \underline{A}daptive online \underline{G}radient desc\underline{E}nt with infeasib\underline{L}e projection). BAGEL constructs a Lyapunov-weighted surrogate that combines the revealed cost and constraint functions, passes the resulting surrogate gradients to a blocked adaptive OCO subroutine, and implements the subroutine's infeasible projections through an SO. The main analytical difficulty is that the surrogate gradients depend on the accumulated constraint violation and can therefore vary substantially over time. Our analysis jointly controls the surrogate regret and the movement of successive iterates. The former produces the regret and CCV guarantees, while the latter connects the online analysis to the number of SO calls required by the infeasible-projection procedure. The main contributions are as follows.
\begin{enumerate}[leftmargin=*,itemsep=2pt,parsep=0pt]

\item
\textbf{Adversarial COCO under separation-oracle access.}
For convex costs and any $\beta\in(0,1/2]$, BAGEL achieves $\C{O}(T^{1-\beta})$ regret and $\C{O}(T^{1-\beta}\log T)$ CCV using $\widetilde{\C{O}}((D/r)^2T^{2\beta})$ SO calls. At $\beta=1/2$, this translates to a $\sqrt{T}$ polynomial horizon dependence regret with a near-linear number of SO calls.

\item
\textbf{Query analysis for Lyapunov-weighted surrogate gradients.}
We analyze an adaptive OCO subroutine whose step sizes respond to the time-varying norms of the COCO surrogate gradients. The technical contribution is the joint control of surrogate regret and iterate movement required to translate the infeasible-projection construction into an SO-query bound, with the added $\epsilon$-stabilized denominator acting as a technical regularization device.

\item
\textbf{An explicit performance-oracle tradeoff.}
The block parameter $\beta$ exposes a tradeoff between regret, cumulative violation, and the number of SO calls. The query guarantee also retains the geometry factor $(D/r)^2$, making explicit the regimes in which SO access may be advantageous. Accordingly, our result characterizes what is achievable under an SO access model rather than asserting that an SO is universally cheaper than a PO or LOO.
As a secondary extension, we analyze strongly convex cost functions in Appendix, which, however, is not part of the paper's primary claim.

\end{enumerate}

\paragraph{Paper Outlook.} The remainder of the paper is organized as follows. Section~\ref{sec:relatedWork} reviews the relevant adversarial COCO and projection-free OCO literature. Section~\ref{sec:prelim} introduces the notation, geometric quantities, auxiliary violation measure, and standing assumptions. Section~\ref{sec:oco} develops the adaptive OCO base algorithm under SO access and establishes its regret and query guarantees. Section~\ref{sec:pacogd} introduces BAGEL and derives its regret, CCV, and SO-query bounds. The strongly convex extensions and complete proofs are provided in the appendices as secondary results.

\section{Literature Review} \label{sec:relatedWork}

\paragraph{Projection-based algorithms for adversarial COCO.}\quad
Early work on online optimization with long-term or adversarial constraints established regret--violation tradeoffs using projection or more general constrained-optimization subroutines \citep{mahdavi2012trading}. For adversarially revealed constraints, \citet{guo2022online} obtained $\C{O}(\sqrt{T})$ regret and $\C{O}(T^{3/4})$ CCV for convex costs by solving a constrained convex optimization problem in each round, which is also covered by the more general framework of \citet{yi2023distributed}. Using a PO together with a Lyapunov-weighted surrogate, \citet{sinha2024optimal} obtained $\C{O}(\sqrt{T})$ regret and $\C{O}(\sqrt{T}\log T)$ CCV for convex costs, as well as a strongly convex extension.
Subsequent PO-based work has refined these guarantees in several directions. \citet{sinha2025beyond} derived dimension-dependent regret-CCV tradeoffs that permit smaller violation at the price of additional regret. \citet{vaze2025instance} retained $\C{O}(\sqrt{T})$ regret while obtaining an instance-dependent violation bound that can improve substantially when the revealed constraint sets have favorable geometry. More recently, \citet{sarkar2026geometric} obtained $\C{O}(\sqrt{T})$ regret and $\C{O}(\sqrt{T})$ CCV for convex costs, and logarithmic regret and CCV for strongly convex costs, using a PO. We include this fixed-horizon result in Table~\ref{tbl}. Separately, \citet{sarkar2026anytime} developed a projection-based anytime algorithm that provides guarantees at every intermediate horizon without knowing $T$ in advance. Because its distinguishing contribution is horizon adaptivity rather than a different fixed-horizon oracle tradeoff, we discuss it here but do not add a separate row to the table.

\begin{table*}[ht]
\begin{center}
\resizebox{\textwidth}{!}{%
\begin{tabular}{lccccccr}
\toprule
Reference & Regret  & CCV  &  $\beta$ Range & Projection & \makecell{Oracle\\ type} & \makecell{Total Calls \\ to Oracle}   & Cost Function    \\
\midrule

\multirow{2}{*}{\citet{guo2022online}} & $\C{O}(T^{1/2})$  &  $\C{O}(T^{3/4})$  & - & \multirow{2}{*}{Based}  &   \multirow{2}{*}{COO} & -   & Convex  \\
 & $\C{O}(\log(T))$  &  $\C{O}(T^{1/2} \sqrt{\log(T)})$  & -  &  &  & $\C{O}(T)$ & Strongly Convex \\
 \midrule
\multirow{2}{*}{\citet{yi2023distributed}}  & $\C{O}(T^{\max(\beta,1-\beta)})$  & $\C{O}(T^{1-\beta/2})$ &  (0,1)  & \multirow{2}{*}{Based} &  \multirow{2}{*}{COO}  & $\C{O}(T)$    & Convex  \\
  & $\C{O}(\log(T))$  &  $\C{O}(T^{1/2} \sqrt{\log(T)})$ & - &  &  & $\C{O}(T)$   & Strongly Convex \\
  \midrule
\multirow{2}{*}{\citet{sinha2024optimal}}   & $\C{O}(T^{1/2})$       & $\C{O}(T^{1/2}\log(T))$ & - & \multirow{2}{*}{Based} & \multirow{2}{*}{PO} & $\C{O}(T)$  & Convex    \\
 & $\C{O}(\log(T))$     & $\C{O}(T^{1/2} \sqrt{\log(T)})$ & - &  &   & $\C{O}(T)$  & Strongly Convex  \\
 \midrule
\multirow{2}{*}{\citet{sarkar2026geometric}} & $\C{O}(T^{1/2})$ & $\C{O}(T^{1/2})$ & - & \multirow{2}{*}{Based} & \multirow{2}{*}{PO} & $\C{O}(T)$ & Convex \\
 & $\C{O}(\log T)$ & $\C{O}(\log T)$ & - & & & $\C{O}(T)$ & Strongly Convex \\
 \midrule
\citet{garber2024projection} (Thm 4.1) & $\C{O}(T^{3/4})$  &  $\C{O}(T^{7/8})$ & - & Free &   LOO & $\C{O}(T)$  & Convex  \\
 \midrule
\citet{sarkar2025projection} (Thm 3.4) & $\C{O}(T^{3/4})$  &  $\C{O}(T^{3/4})$ & -  & Free & LOO   & $\C{O}(T)$ & Convex  \\
 \midrule
\citet{wang2025revisiting} & $\mathcal{O}(T^{3/4})$ & $\mathcal{O}(T^{3/4} \log T)$ & - & Free & LOO & $\mathcal{O}(T)$ & Convex \\
 \midrule
\multirow{2}{*}{This paper}  & $\C{O}({T}^{1-\beta})$       & $\C{O}({T}^{1-\beta}\log(T))$ & (0,1/2] & \multirow{2}{*}{Free} &  SO  & $\tilde{\C{O}}(T^{2\beta})$  & Convex     \\
& $\C{O}(T^{1-\beta}\log(T))$       & $\C{O}(T^{1-\beta/2}\sqrt{\log(T)})$ & (0,1] &  &  SO  & $\tilde{\C{O}}(T^{2\beta})$  & Strongly Convex     \\
\bottomrule
\end{tabular}
}
\end{center}
\caption{
Summary of fixed-horizon guarantees for adversarial COCO, showing dependence on the horizon $T$. Projection-based methods access the action set through a convex optimization oracle (COO) or projection oracle (PO), whereas projection-free methods use a linear optimization oracle (LOO) or separation oracle (SO). The table suppresses factors independent of $T$. In particular, BAGEL's SO-query bound contains the geometry factor $(D/r)^2$. For convex costs, choosing $\beta=1/2$ gives $\C{O}(\sqrt{T})$ regret and $\C{O}(\sqrt{T}\log T)$ CCV using $\widetilde{\C{O}}(T)$ SO calls, up to geometry-dependent factors. BAGEL's detailed guarantees are in Theorem~\ref{thm:main} for convex costs and Theorem~\ref{thm:constrained-strongly} for strongly convex costs.
}
\label{tbl}
\end{table*}

\paragraph{Projection-free Algorithms for OCO} \quad Many projection-free algorithms have been proposed for classical OCO. Frank--Wolfe approaches achieve $\C{O}(T^{3/4})$ regret for convex costs \cite{pmlr-v37-daniely15,hazan2012projection} and $\C{O}(T^{2/3})$ regret under additional curvature assumptions \cite{kretzu2021revisiting,wan2021projection}. \citet{garber2021efficient} introduced infeasible projection as a replacement for exact Euclidean projection in online linear optimization. In parallel with \citet{mhammedi2022efficient}, \citet{garber2022new} implemented infeasible projection through an SO for OCO and obtained $\C{O}(T^{1/2})$ regret using an online-gradient method with a constant step size; their LOO implementation gives a different regret--query tradeoff. Recent work by \citet{mhammedi2025online} further develops SO access for standard online convex optimization.
Our algorithms use the SO-based infeasible-projection framework of \citet{garber2022new}, specifically the variant of \citet{pedramfar2024linearizable} that returns actions in $\K$. 

\paragraph{Projection-free Algorithms for Adversarial COCO} \quad
Projection-free methods for adversarial COCO have primarily assumed access to an LOO. \citet{garber2024projection} obtained $\C{O}(T^{3/4})$ regret and $\C{O}(T^{7/8})$ CCV, while \citet{sarkar2025projection} improved the violation guarantee and obtained $\widetilde{\C{O}}(T^{3/4})$ for both regret and CCV. \citet{wang2025revisiting} obtained $\C{O}(T^{3/4})$ regret and $\C{O}(T^{3/4}\log T)$ CCV. These results and BAGEL operate under different action-set access models: the former operate under LOO access, whereas BAGEL works for SO access setting. Consequently, comparisons between them describe regret-violation-query tradeoffs under different oracle assumptions rather than a universal ordering of their computational costs.

BAGEL combines components originating in these different lines of research. Our Lyapunov-weighted surrogate is inspired by the projection-based adversarial-COCO analysis of \citet{sinha2024optimal} and the infeasible-projection mechanism follows \citet{garber2022new} and \citet{pedramfar2024linearizable}. The contribution is the analysis that makes these components compatible under adversarial constraints. In particular, the surrogate gradients depend on accumulated violation and may vary substantially over time, while the number of SO calls depends on the movement of the resulting iterates. Our analysis jointly controls regret, cumulative violation, and this iterate movement, producing an explicit SO-query bound with its dependence on the geometry of $\K$, with the added $\epsilon$-stablized denominator acting as an technical regularization.


\section{Preliminaries}
\label{sec:prelim}

\paragraph{Notation and function classes.}
For a positive integer $m$, let $[m]\triangleq\{1,\ldots,m\}$.
We use $\langle\cdot,\cdot\rangle$ and $\|\cdot\|$ to denote the Euclidean inner product and norm, respectively, and define $(z)^+\triangleq\max\{z,0\}$. 
For a convex function $h:\K\to\mathbb{R}$, let $\partial h(\x)$ denote its subgradient at $\x$. Whenever we write $\nabla h(\x)$ for a non-differentiable convex function, it denotes an arbitrary element of
$\partial h(\x)$.
A function $h:\K\to\mathbb{R}$ is convex if, for every
$\x,\y\in\K$ and every $\mathbf{s}\in\partial h(\x)$, $h(\y) \geq h(\x)+\langle \mathbf{s},\y-\x\rangle.$
We say it is $\theta$-strongly convex for some $\theta>0$ if $h(\y) \geq h(\x)+\langle \mathbf{s},\y-\x\rangle +\frac{\theta}{2}\|\y-\x\|^2$ for every $\x,\y\in\K$ and every $\mathbf{s}\in\partial h(\x)$.

\paragraph{Constraint aggregation.}
At round $t$, define the maximum constraint function $g_t(\x)\triangleq\max_{i\in[k]}g_{t,i}(\x).$
Because a pointwise maximum of convex functions is convex, $g_t$ is
convex, and $(g_t(\x))^+ = \max_{i\in[k]}\bigl(g_{t,i}(\x)\bigr)^+.$
In the analysis, we control the stronger roundwise-max violation
\[
    \mathtt{V}(T) \triangleq \sum_{t=1}^T\bigl(g_t(\x_t)\bigr)^+ =\sum_{t=1}^T \max_{i\in[k]}\bigl(g_{t,i}(\x_t)\bigr)^+.
\]
The standard cumulative constraint violation defined in the
Introduction satisfies
\[
    \mathtt{CCV}(T)
    =
    \max_{i\in[k]}
    \sum_{t=1}^T\bigl(g_{t,i}(\x_t)\bigr)^+
    \leq
    \mathtt{V}(T).
\]
Thus, every upper bound established for $\mathtt{V}(T)$ also yields
the same upper bound for $\mathtt{CCV}(T)$.

\paragraph{Action-set geometry.}
For a nonempty convex set $\K\subseteq\mathbb{R}^d$, let $\operatorname{aff}(\K)$ and $\operatorname{relint}(\K)$ denote its affine hull and relative interior, respectively. The diameter of $\K$ is $D\triangleq \max_{\x,\y\in\K}\|\x-\y\|.$
We assume that a point $\mathbf{c}\in\operatorname{relint}(\K)$ and a certified radius $r>0$ are known such that $\mathbb{B}_r(\mathbf{c}) \cap\operatorname{aff}(\K) \subseteq \K,$ where $\mathbb{B}_r(\mathbf{c})$ is the Euclidean ball of radius $r$ centered at $\mathbf{c}$. The radius $r$ need not be the largest radius satisfying this inclusion.

For a dimensionless shrinking parameter $\delta\in(0,1)$, define the shrunk set $\K_\delta \triangleq (1-\delta)\K+\delta\mathbf{c} = \{ (1-\delta)\x+\delta\mathbf{c}:\x\in\K \}.$
Convexity of $\K$ implies $\K_\delta\subseteq\K$. In addition, for every $\x\in\K$, the corresponding point $\widetilde{\x} \triangleq (1-\delta)\x+\delta\mathbf{c} \in\K_\delta$ satisfies $\|\widetilde{\x}-\x\| = \delta\|\mathbf{c}-\x\| \leq \delta D.$

\paragraph{Assumptions.} Under the given problem setting, we introduce a few assumptions. The first one is about action set and the access model, the second one regularizes the objective functions, and the third one ensure feasibility of the problem.

\begin{assumption}[Action set and access model]
\label{asm:action-set}
The action set $\K\subseteq\mathbb{R}^d$ is nonempty, convex, closed, and bounded, with diameter at most $D$. The learner is given $\mathbf{c}\in\operatorname{relint}(\K)$ and $r>0$ satisfying $\mathbb{B}_r(\mathbf{c}) \cap\operatorname{aff}(\K) \subseteq\K,$ and has access to an exact separation oracle for $\K$.

If $\K$ is not full-dimensional, we additionally assume that a
representation of $\operatorname{aff}(\K)$ is available and permits
orthogonal projection onto $\operatorname{aff}(\K)$ and its associated
linear subspace. These affine-subspace projections are not counted as
separation-oracle calls.
\end{assumption}

\begin{assumption}[Convex and Continuous Functions]
\label{asm:functions}
$\forall t\in[T], \forall i\in[k]$, the cost function $f_t$ and constraint function $g_{t,i}$ are convex on $\K$. Their subgradients are uniformly bounded, i.e., $\forall\x\in\K$, $\|\mathbf{s}_f\|\leq M_1$ and $\|\mathbf{s}_{g,i}\|\leq M_1$ for all $\mathbf{s}_f\in\partial f_t(\x)$ and $\mathbf{s}_{g,i}\in\partial g_{t,i}(\x)$. Thus, $f_t$ and $g_t=\max_{i\in[k]}g_{t,i}$ are $M_1$-Lipschitz continuous over $\K$.

For the strongly convex extension, each $f_t$ is additionally assumed
to be $\theta$-strongly convex for a common parameter $\theta>0$ while
the constraint functions remain convex.
\end{assumption}

\begin{assumption}[Common feasibility]
\label{asm:feas}
There exists an action $\x^\star\in\K$ satisfying
\[
    g_{t,i}(\x^\star)\leq 0
    \qquad
    \forall t\in[T], \ \forall i\in[k].
\]
Equivalently, the common feasible comparator set $\K^\star
    \triangleq
    \left\{
        \x\in\K:
        g_{t,i}(\x)\leq 0, 
        \ \forall t\in[T],\ \forall i\in[k]
    \right\}$ is nonempty.
\end{assumption}

\paragraph{Separation-oracle access.}
Given a query point $\y\in\mathbb{R}^d$, an exact separation oracle for $\K$, denoted by $\operatorname{SO}_{\K}$, either certifies that $\y\in\K$ or returns a nonzero vector $\mathbf{a}$ such that $\forall \x\in\K$, $\langle\mathbf{a},\y-\x\rangle>0$.
Our oracle-complexity results count calls to $\operatorname{SO}_{\K}$. They do not count elementary arithmetic, first-order evaluations of the revealed functions, or the affine-subspace projections specified in Assumption~\ref{asm:action-set}.

\paragraph{Infeasible projection through an SO.}
We use the Infeasible Projection via Separation Oracle
(IP-SO) procedure of \citet{pedramfar2024linearizable}, building on
the infeasible-projection framework of \citet{garber2022new}.
Given a point $\y_0\in\mathbb{R}^d$, the procedure returns a feasible
point $\y\in\K$ satisfying
\[
    \|\y-\x\|
    \leq
    \|\y_0-\x\|,
    \qquad
    \forall \x\in\K_\delta.
\]
Thus, although $\y$ need not be the Euclidean projection of $\y_0$
onto $\K$, it does not increase the distance from $\y_0$ to any
point in the shrunk set $\K_\delta$.

We denote this procedure by
$\mathcal{P}_{\K,\mathbf{c},r,\delta}$, or simply by $\mathcal{P}$
when its parameters are clear. Its implementation and
separation-oracle complexity are stated in
Algorithm~\ref{alg:ipso} and Lemma~\ref{lem:ipso} in
Appendix~\ref{apx:ipso}.

\section{Base Algorithm: OCO with Separation-Oracle Access}
\label{sec:oco}

In order to extend OCO methods to the COCO framework inspired by \citet{sinha2024optimal}, we use a surrogate function that combines the cost and constraint, leveraging the cumulative constraint violation as weights. 
A key challenge is that the gradients can have vastly different magnitudes across iterations, as shown in Section~\ref{sec:pacogd}. When constraint violations accumulate, the weight can become very large, leading to explosive gradients. An adaptive step-size dampens these large gradients by incorporating gradient magnitudes into the denominator, providing the stability required for the surrogate-based analysis to hold. Therefore, we first formally analyze a projection-free OGD with such an adaptive step-size (Algorithm~\ref{alg:general}), a necessary building block for the algorithm under adversarial constraints.

The proposed base algorithm is based on Online Gradient Descent with adaptive step-size. Although \citet{garber2022new} proposed a projection free OGD using an infeasible projection oracle (Algorithm~1, Lemma~4), it only considered constant step size and did not consider feasibility constraints on the action set $\K$. 
However, the guarantees of results for COCO algorithm require adaptive step-sizes and feasible solutions returned by infeasible projection oracle, which is why we propose and analyze the Base Algorithm described in Algorithm~\ref{alg:general}. 
Note that we add a constraint constant $\epsilon$ so that the regret are guaranteed with only $\tilde{\C{O}}(T)$ total calls to the separation oracle.

Algorithm~\ref{alg:general} takes as input: action set $\K$, time horizon $T$, $\BF{c} \in \op{relint}(\K)$, $r = r_{\K, \BF{c}}$, block size $K$, shrinking parameter $\delta\in[0,1)$, diameter of the action set $D$, an instance of IP-SO algorithm $\C{P}$ implemented by Algorithm~\ref{alg:ipso}, a constraint constant $\epsilon>0$ for convex cost functions or the strongly-convex parameter $\theta$ for strongly-convex cost functions.
We initiate the program by choosing $\x_1\in\K$, and pass $\K, \BF{c}, r, \delta$ to $\C{P}$ to initiate algorithm instances.
Within each block $m$, there are $K$ iterations, represented by $\C{T}_m$.
At each step $t$, we play action $\x_t$, and observe the cost function $f_t$ adversarially selected and revealed by the adversary, which we use to compute $\nabla_t$, the gradient of $f_t$ at $\x_t$.
After $K$ iterations, we compute the average gradient in the block $m$, $\bar{\nabla}_m=\frac{1}{K} \sum_{t\in\C{T}_m} \nabla_t $, and perform a gradient descent update, $\x_m - \eta_m\bar{\nabla}_m$, with adaptive step-size $\eta_m = \frac{D}{\sqrt{\epsilon + \sum^m_{\tau=1}\Vert \bar{\nabla}_\tau \Vert^2}}$ in the convex case, or $\eta_m = \frac{1}{m \theta}$ in the strongly convex case.
The gradient descent update $\x_m-\eta_m\bn_m$ is passed to the IP-SO instance $\C{P}$ to obtain next action, $\x_{m+1}$.

\begin{algorithm}[H] 
\caption{Base Algorithm}
\label{alg:general}
\begin{algorithmic}[1]
  \Require  Action set $\K$, time horizon $T$, block size $K$, $\BF{c} \in \op{relint}(\K)$, $r = r_{\K, \BF{c}}$, shrinking parameter $\delta\in[0,1)$, diameter of the action set $D$, an instance of IP-SO algorithm $\C{P}$ implemented by Algorithm~\ref{alg:ipso}, (in the convex case) constant $\epsilon > 0$, (in the strongly-convex case) strongly-convex parameter $\theta$
  \State \textbf{Initialization:} Pick $\mathbf{x}_1\in\mathcal{K}$; pass $\mathcal{K},\mathbf{c},r,\delta$ to $\mathcal{P}$
  \For{$m \in [T/K]$}
    \For{$t\in\mathcal{T}_m$}
      \State Play $\mathbf{x}_m$
      \State Adversary reveals $f_t$
      \State Compute $\nabla_t \gets \nabla f_t(\mathbf{x}_m)$
    \EndFor
    \State $\bar{\nabla}_m \gets \frac{1}{K}\sum_{t\in\mathcal{T}_m}\nabla_t$
    \State \textbf{Step size:}
    \Statex \hspace{\algorithmicindent}$\eta_m \gets \frac{D}{\sqrt{\epsilon + \sum_{\tau=1}^m \|\bar{\nabla}_\tau\|^2}}$ (Convex)
    \Statex \hspace{\algorithmicindent}$\eta_m \gets \frac{1}{m\,\theta}$ (Strongly Convex)
    \State $\x_{m+1}\gets \C{P}(\x_m - \eta_m\bn_m)$
  \EndFor
\end{algorithmic}
\end{algorithm}

\begin{theorem}\label{thm:unconstrained-regret}
Assume functions $f_t$ are convex and $G_t$-Lipschitz continuous. Let $\mathbf{x}_0^* \in \arg\min_{\mathbf{x} \in \mathcal{K}^*} \sum_{t=1}^T f_t(\mathbf{x})$. Then, Algorithm~\ref{alg:general} ensures that the regret against the sequence minimizer $\mathbf{x}_0^*$\footnote{
Since Algorithm~\ref{alg:general} does not consider adversarial constraint, $\K=\K^*$ for Theorem~\ref{thm:unconstrained-regret}. To avoid confusion, we use $\x^*_0$ when we do not consider adversarial constraints, and $\x^*$ when we do.
} is bounded as
\begin{small}
\begin{align*}
\mathtt{Regret}(T) &= \sum_{t=1}^T f_t(\x_t) - \sum_{t=1}^T f_t(\mathbf{x}_0^*) \\
&\leq \frac{3}{2} D K \sqrt{\sum_{m=1}^{T/K} \Vert \bar{\nabla}_m \Vert^2} + D \delta \sum_{t=1}^T G_t + \frac{KD}{2}\sqrt{\epsilon},
\end{align*}
\end{small}
where $\delta \in [0,1)$ is the shrinking parameter, $r = r_{\mathcal{K},\mathbf{c}}$, $K$ is the block size, $D$ is the diameter of $\mathcal{K}$, and the definition of $\bar{\nabla}_m$ is given in Algorithm~\ref{alg:general} (line 9). If $\epsilon > 0$ is a constant independent of $T$, then this is achieved with $\mathcal{O}\left((D/r)^2\delta^{-2}\log(T) + T K^{-1}\right)$ calls to a separation oracle. On the other hand, if we only assume $\epsilon \geq 0$, then the same bounds for regret and CCV are achieved with $\mathcal{O}\left((D/r)^2T K^{-1}\delta^{-2}\right)$ calls to a separation oracle.
\end{theorem}

\begin{remark}\label{rmk:1}
    When block size $K$ is equal to $1$ and step-size is constant, this result reduces to a variant of Theorem~14 in~\citet{garber2022new}.
    Also note that the aforementioned lemma allows for the the actions to be outside the constraint set, which is why their bound does not include the term $ D \delta  \sum_{t=1}^T G_t$.
\end{remark}

\begin{proof}[Proof Sketch]
We introduce a shrunk optimal action $\tilde{\mathbf{x}}_0^* \in \mathcal{K}_\delta$ as a bridge to decompose the total regret into two bounds: the regret against $\tilde{\mathbf{x}}_0^*$ and the residual distance to the true sequence minimizer $\mathbf{x}_0^*$. Using the distance-reducing properties of the IP-SO algorithm (Lemma~\ref{lem:ipso}), the first term is bounded by a summation involving our adaptive step sizes and squared gradient norms. Applying standard integral bounds to this summation yields the desired $\mathcal{O}(T^{1/2})$ regret. On the other hand, bounding the total separation oracle calls relies on tracking the movement between consecutive iterates, which produces the sum $\sum_{m=1}^{T/K} \eta_m^2 \|\bar{\nabla}_m\|^2$. Integrating this $\epsilon$-stabilized sequence yields a $\log\left(1 + \frac{M_1 T}{\epsilon K}\right)$ term, explicitly demonstrating why $\epsilon > 0$ is mathematically required to prevent the oracle complexity bound from diverging to infinity. \emph{(See Appendix~\ref{apx:unconstrained-regret} for the complete proof).}
\begin{remark} \label{rmk:epsilon}
        If we only assume that $\epsilon \geq 0$, instead of $\epsilon>0$, then the above regret bound becomes infinite as it contains the term $\log\left( 1 + \frac{M_1 T}{\epsilon K} \right)$.
        However, we may instead bound $\sum_{m = 1}^{T/K} \frac{\| \bn_m \|^2}{\epsilon + \sum_{\tau=1}^{m}\| \bn_m \|^2}$ with $T/K$, since each term is bounded by $1$. Thus, with Cauchy-Schwartz inequality, we see that $\left( \sum_{m = 1}^{T/K} \eta_m \| \bn_m \| \right)^2 \leq \frac{2D^2 T}{K} \cdot \frac{T}{K}$ and therefore the total number of calls to the separation oracle would be bounded by
        \begin{align*}
         &\frac{2D^2}{\delta^2r^2} \frac{T}{K} + \frac{2 \sqrt{2} D^2}{r^2 \delta} \frac{T}{K} + \frac{D^2}{r^2} + \frac{T}{K} 
        = \C{O}\left( (D/r)^2 T K^{-1} \delta^{-2} \right).
         \qedhere
        \end{align*}
    \end{remark}
\end{proof}

\begin{corollary}
\label{cor:con}
In Theorem~\ref{thm:unconstrained-regret}, if we further assume that functions are $M_1$-Lipschitz continuous, then we see that
\begin{align*}
\mathtt{Regret}(T) & \leq \frac{3}{2} D \sqrt{M_1 K T} + D \delta T M_1 + \frac{KD}{2}\sqrt{\epsilon}.
\end{align*}
In particular, if $K = \Theta(T^{1 - 2\beta})$ and $\delta = \Theta(T^{-\beta})$ for some $\beta \in [0, 1/2]$, then $\mathtt{Regret}(T) = \C{O}(DM_1^{1/2}T^{1 - \beta})$ with $\tilde{\C{O}}(T^{2\beta})$ calls to the separation oracle.
On the other hand, if we assume that $\epsilon = 0$, we obtain the same bounds for regret and CCV with $\tilde{\C{O}}(T^{4\beta})$ calls to the separation oracle.
\end{corollary}

We note that better regret bounds are achievable when the cost functions are chosen to be strongly convex. 
Thus, we give Theorem~\ref{thm:unconstrained-strongly} to describe results for strongly convex functions. 
The proof details for Theorem~\ref{thm:unconstrained-strongly} and Corollary~\ref{cor:strcon} can be found in Appendix~\ref{apx:unconstrained-strongly}.

\section{BAGEL for Adversarial COCO under Separation-Oracle Access}\label{sec:pacogd}

In this section, to solve Online Convex Optimization with Adversarial Constraint (Adversarial COCO), we propose Algorithm~\ref{alg:constrained}, \underline{B}locked \underline{A}daptive online \underline{G}radient d\underline{E}scent with infeasib\underline{L}e projection ($\mathtt{BAGEL}$), where ``infeasible projection'' refers to the use of IP-SO described in Section~\ref{sec:prelim}. 
For computational convenience, we introduce a processing parameter $\gamma$ and we let $\tf_t = \gamma f_t$, $\tilde{g}_{t,i}=\gamma(g_{t,i})^+$. To track the cumulative constraint violation, we let $Q_t=Q_{t-1}+\tilde{g}_t(\x_t)$, and $Q_0=0$.
Inspired by \citet{sinha2024optimal}, $\mathtt{BAGEL}$ (Algorithm 2) combines costs and constraints into a single surrogate cost function, $\hf_t$, and applies Base Algorithm to it. This surrogate consists of the original cost function and the current constraint violation weighted by dynamic term. If violations have been low, the weight is small, and the algorithm focuses on minimizing cost. If violations have been high, the weight becomes large, forcing the algorithm to aggressively prioritize satisfying the constraint. 
Since our objective is to make the weighted cumulative constraint violation small, similar to \citet{sinha2024optimal}, we introduce a potential function $\Phi: \BB{R}^+\mapsto\BB{R}^+$ that is non-decreasing, differentiable, convex, and satisfies $\Phi(0)=0$. Thus, we create the surrogate cost function to be $\hf = \tf_t + \Phi'(Q_t)\tg_t$ where $\Phi'(Q_t)$ is the gradient of the pre-defined function $\Phi$ at $Q_t$.

\begin{algorithm}[H] 
\caption{\textsc{BAGEL}(\underline{B}locked \underline{A}daptive online \underline{G}radient desc\underline{E}nt with infeasib\underline{L}e projection)}
\label{alg:constrained}
\begin{algorithmic}[1]
  \Require Action set $\K$, time horizon $T$, $\BF{c} \in \op{relint}(\K)$, $r = r_{\K, \BF{c}}$, shrinking parameter $\delta \in [0,1)$, block size $K$, action set diameter $D$, an instance of the Base Algorithm (Algorithm~\ref{alg:general}) $\C{A}$, processing parameter $\gamma$, regularization parameter $V$, Lyapunov function $\Phi(\cdot)$, an instance of IP-SO (Algorithm~\ref{alg:ipso}) $\C{P}$, (in the convex case) constant $\epsilon > 0$, (in the strongly-convex case) strongly-convex parameter $\theta$. 
  \State \textbf{Initialization:} Pick $\mathbf{x}_1\in\mathcal{K}$; set $Q_0\gets 0$; pass $\mathcal{K},T,K,\mathbf{c},r,\delta,D,\mathcal{P}$, and $(\epsilon\text{ or }\theta)$ to $\mathcal{A}$
  \For{$m \in [T/K]$}
    \For{$t\in\mathcal{T}_m$}
      \State Play $\mathbf{x}_m$
      \State Adversary reveals $f_t$ and $g_t$
      \State Observe $f_t(\mathbf{x}_m)$ and $(g_t(\mathbf{x}_m))^{+}$
      \State Let $\tilde f_t \gets \gamma\, f_t$\;; \quad $\tilde g_t \gets \gamma\, (g_t)^{+}$
      \State Update cumulated violation $Q_t \gets Q_{t-1} + \tilde g_t$
      \State Calculate gradient of the surrogate objective $\hat f_t \gets V\,\tilde f_t + \Phi'(Q_t)\,\tilde g_t$\;; \quad $\nabla_t \gets \nabla \hat f_t(\mathbf{x}_m)$
      \State Pass $\nabla_t$ to $\mathcal{A}$
    \EndFor
    \State Receive $\mathbf{x}_{m+1}$ from $\mathcal{A}$
  \EndFor
\end{algorithmic}
\end{algorithm}

The algorithm use the inputs: action set $\K$, time horizon $T$, $\BF{c} \in \op{relint}(\K)$, $r = r_{\K, \BF{c}}$, shrinking parameter $\delta \in [0,1)$, block size $K$, action set diameter $D$, an instance of the Base Algorithm (Algorithm~\ref{alg:general}) $\C{A}$, processing parameter $\gamma$, function $\Phi(\cdot)$, an instance of IP-SO $\C{P}$, and constant $\epsilon > 0$ for convex costs or strongly-convex parameter $\theta$. 
We initiate the algorithm by picking the initial action  $\x_1\in\K$, and set $Q_0=0$, and passing the required input to $\C{A}$ to initiate the algorithm instance. 
We break the horizon $T$ into blocks with size $K$. 
For each step $t$ within block $m$, represented by $\C{T}_m$, we play action $\x_m$ determined by the algorithm, and then observe cost $f_t(\x_t)$ and constraint violation $(g_t(\x_t))^+$, where the cost function and the maximum constraint function are chosen and revealed by an adversary. 
We transform them into $\tilde{f}_t = \gamma f, \tilde{g}_t =\gamma (g_t(x_t))^+$, and update the cumulative constraint violation with $Q_t = Q_{t-1}+\tilde{g}_t$. 
We compute the gradient $\nabla_t$ of the surrogate function $\hat{f}_t(x_t):=\tilde{f}_t + \Phi'(Q_t)\tilde{g}_t$, i.e., $\nabla_t=\nabla \hft$, and pass the gradient information to $\C{A}$.
Because the operator $(\cdot)^+$ introduces a point of non-differentiability at zero, any gradients evaluated for $\tilde{g}_t$ and the resulting surrogate $\hat{f}_t$ are technically subgradients. We note that standard online gradient descent and projection-free guarantees apply identically to subgradients, preserving all theoretical bounds.
At the end of the each block, $\C{A}$ returns the action for the next block $\x_{m+1}$. 
The following Theorem provides the regret and the CCV for the proposed algorithm in the convex case.

\begin{theorem}
\label{thm:main}
    Assume cost functions $f_t$ and constraint functions $g_t$ are convex and $M_1$-Lipschitz continuous over action set $\K$, with diameter $D$.
    Let $\BF{c} \in \op{relint}(\K)$, and $r = r_{\K, \BF{c}}$.
    With $\beta\in[0,\frac{1}{2}]$, if we choose $\gamma=(M_1R)^{-1}$, $\delta=\Theta(T^{-\beta})$, $K= \Theta(T^{1-2\beta})$, and $\Phi(x)=e^{\lambda x}-1$ where $\lambda=(2\delta T + 3\sqrt{2TK})^{-1}$,
    Algorithm~\ref{alg:constrained} ensures that:
    \begin{align*}
        \mathtt{Regret}(T) 
        &\leq D M_1 \delta T + \frac{3}{\sqrt{2}} DM_1\sqrt{TK} + DM_1 + \frac{K D^2 M_1}{2}\sqrt{\epsilon}
        = \C{O}(DM_1T^{1-\beta}),
    \end{align*}
    and 
    \begin{align*}
        \mathtt{CCV}(T) 
        &\leq 2 D M_1 (\delta T + \frac{3}{\sqrt{2}}\sqrt{TK}) \C{O}(\log T) 
        = \C{O}(DM_1T^{1-\beta}\log T).
    \end{align*}    
    with $\tilde{\C{O}}((D/r)^2T^{2\beta})$ calls to the separation oracles. 
\end{theorem}

\begin{proof}[Proof Sketch]
The proof proceeds by analyzing the performance of the base OCO algorithm (Theorem 1) on the sequence of surrogate costs $\hat{f}_t$. First, we bound the surrogate regret $\mathcal{R}(\hat{f})$, noting that the surrogate gradients are amplified by the Lyapunov function's derivative $\Phi'(Q_t)$, which captures the worst-case constraint violation. We then establish a critical bridge inequality, $\mathcal{R}(\hat{f}) \geq \mathcal{R}(\tilde{f}) + \Phi(Q_T)$, leveraging the convexity of $\Phi$ to explicitly link the boundable surrogate regret to both the true regret and the cumulative constraint violations. Finally, by selecting an exponential Lyapunov function $\Phi(Q_T) = e^{\lambda Q_T} - 1$ and tuning the hyperparameters ($\lambda, \gamma, K, \delta$), we offset the penalty terms to achieve the final $\mathcal{O}(T^{1-\beta})$ regret and $\mathcal{O}(T^{1-\beta}\log T)$ CCV bounds. \emph{(See Appendix~\ref{apx:constrained-main} for the complete derivation).}
\end{proof}

\begin{remark}[Performance-Oracle Tradeoff]
Note that if we take $\beta=\frac{1}{2}$, we match the $\sqrt{T}$ for both regret and CCV of projection-based methods, with $\tilde{\C{O}}(T)$ total calls to the separation oracles. In this case, there will be no blocking effect because block size $K=1$. 
However, considering the diverse application scenario, we introduce the trade-off parameter $\beta$ to allow adjustment of the algorithm to real-life computational limitations.
When $\beta=\frac{1}{4}$, we have $\C{O}(T^{3/4})$ result for regret and CCV, which matches \citet{garber2024projection} on regret and \citet{sarkar2025projection} on regret and CCV, with at most $\tilde{\C{O}}(\sqrt{T})$ total calls to the separation oracles.
\end{remark}

\begin{remark}[Set Geometry Dependence of SO Complexity]
Our total calls to the SO contains the parameter $(D/r)^2$, which represents the condition number of the constraint set geometry (the ratio between the diameter and the radius of the largest enclosed ball). For `well-conditioned' sets (e.g., a ball or hypercube), this ratio is small or constant. However, for `ill-conditioned' sets in high dimensions, $D/r$ can scale polynomially with the dimension $d$ (e.g., $\sqrt{d}$ or $d$). Thus, BAGEL offers a significant advantage in regimes where the time horizon $T$ is the dominant factor ($T \gg (D/r)^2$), whereas LOO methods remain preferable for high-dimensional sets where $D/r$ is prohibitively large.
\end{remark}

\section{Conclusions}

 In this paper, we investigated the adversarially constrained online convex optimization under separation oracle access to the underlying action set.
$\mathtt{BAGEL}$ is the first projection-free adversarial COCO algorithm under this specific access setting, and we achieve $\tilde{\C{O}}(T^{1/2})$ regret and CCV bounds of projection-based methods, using a number of separation oracle calls that is near-linear in the time horizon $T$.

\paragraph{Limitations.} The practical efficiency of a Separation Oracle is not universal, and it depends heavily on the underlying geometry of the action set. For example, in robust control and generative adversarial networks, constraints are often modeled as spectral norm balls \cite{miyato2018spectral, zhou1996robust}, where an SO is highly efficient, requiring only a computationally cheap rank-one SVD. On the other hand, in tasks like collaborative filtering, constraints are often modeled as nuclear norm balls \cite{scarvelis2024nuclear}. For these geometries, an SO requires a full SVD, rendering it computationally prohibitive (whereas a LOO would be efficient). As a result, \texttt{BAGEL} is most advantageous in applications where the constraint set geometry naturally favors an SO. 


\newpage
\bibliography{main}

@article{redeker2018portfolio,
  title={Portfolio optimization under dynamic risk constraints: Continuous vs. discrete time trading},
  author={Redeker, Imke and Wunderlich, Ralf},
  journal={Statistics \& Risk Modeling},
  volume={35},
  number={1-2},
  pages={1--21},
  year={2018}
}

@article{doostmohammadian2022distributed,
  title={Distributed anytime-feasible resource allocation subject to heterogeneous time-varying delays},
  author={Doostmohammadian, Mohammadreza and Aghasi, Alireza and Rikos, Apostolos I and Grammenos, Andreas and Kalyvianaki, Evangelia and Hadjicostis, Christoforos N and Johansson, Karl H and Charalambous, Themistoklis},
  journal={IEEE Open Journal of Control Systems},
  volume={1},
  pages={255--267},
  year={2022},
  publisher={IEEE}
}

@article{li2024dynamic,
  title={Dynamic pricing with external information and inventory constraint},
  author={Li, Xiaocheng and Zheng, Zeyu},
  journal={Management Science},
  volume={70},
  number={9},
  pages={5985--6001},
  year={2024},
  publisher={INFORMS}
}

@article{da2019collision,
  title={Collision-free encoding for chance-constrained nonconvex path planning},
  author={da Silva Arantes, Marcio and Toledo, Claudio Fabiano Motta and Williams, Brian Charles and Ono, Masahiro},
  journal={IEEE Transactions on Robotics},
  volume={35},
  number={2},
  pages={433--448},
  year={2019},
  publisher={IEEE}
}

@inproceedings{garber2024projection,
  title={Projection-Free Online Convex Optimization with Time-Varying Constraints},
  author={Garber, Dan and Kretzu, Ben},
  booktitle={Forty-first International Conference on Machine Learning},
  year={2024}
}

@article{shalev2012online,
  title={Online learning and online convex optimization},
  author={Shalev-Shwartz, Shai and others},
  journal={Foundations and Trends{\textregistered} in Machine Learning},
  volume={4},
  number={2},
  pages={107--194},
  year={2012},
  publisher={Now Publishers, Inc.}
}

@inproceedings{garber2022new,
  title={New projection-free algorithms for online convex optimization with adaptive regret guarantees},
  author={Garber, Dan and Kretzu, Ben},
  booktitle={Conference on Learning Theory},
  pages={2326--2359},
  year={2022},
  organization={PMLR}
}

@inproceedings{hazan2012projection,
  author    = {Elad E. Hazan and Satyen Kale},
  title     = {Projection-free online learning},
  booktitle = {Proceedings of the 29th International Conference on Machine Learning (ICML)},
  year      = {2012},
  pages     = {521--528}
}

@article{mahdavi2012trading,
  title={Trading regret for efficiency: online convex optimization with long term constraints},
  author={Mahdavi, Mehrdad and Jin, Rong and Yang, Tianbao},
  journal={The Journal of Machine Learning Research},
  volume={13},
  number={1},
  pages={2503--2528},
  year={2012},
  publisher={JMLR. org}
}

@inproceedings{sinha2024optimal,
  title={Optimal Algorithms for Online Convex Optimization with Adversarial Constraints},
  author={Sinha, Abhishek and Vaze, Rahul},
  booktitle={Advances in Neural Information Processing Systems},
  year={2024}
}

@inproceedings{sinha2025beyond,
  title={Beyond $\widetilde{O}(\sqrt{T})$ Constraint Violation for Online Convex Optimization with Adversarial Constraints},
  author={Sinha, Abhishek and Vaze, Rahul},
  booktitle={Advances in Neural Information Processing Systems},
  volume={38},
  year={2025},
  doi={10.52202/085713-4773}
}

@inproceedings{vaze2025instance,
  title={$O(\sqrt{T})$ Static Regret and Instance Dependent Constraint Violation for Constrained Online Convex Optimization},
  author={Vaze, Rahul and Sinha, Abhishek},
  booktitle={Advances in Neural Information Processing Systems},
  volume={38},
  year={2025},
  doi={10.52202/085713-2059}
}

@article{sarkar2026geometric,
  title={A Geometric Approach to Constrained Online Learning},
  author={Sarkar, Dhruv and Sinha, Abhishek},
  journal={arXiv preprint arXiv:2605.21107},
  year={2026},
  url={https://arxiv.org/abs/2605.21107}
}

@inproceedings{sarkar2026anytime,
  title={Optimal Anytime Algorithms for Online Convex Optimization with Adversarial Constraints},
  author={Sarkar, Dhruv and Sinha, Abhishek},
  booktitle={Proceedings of the 43rd International Conference on Machine Learning},
  year={2026},
  url={https://arxiv.org/abs/2510.22579}
}

@inproceedings{pedramfar2024linearizable,
  title = {From Linear to Linearizable Optimization: A Novel Framework with Applications to Stationary and Non-stationary DR-submodular Optimization},
  author = {Mohammad Pedramfar and Vaneet Aggarwal},
  booktitle = {Advances in Neural Information Processing Systems},
  year = {2024}
}

@article{yi2023distributed,
  title={Distributed Online Convex Optimization with Adversarial Constraints: Reduced Cumulative Constraint Violation Bounds under Slater's Condition},
  author={Yi, Xinlei and Li, Xiuxian and Yang, Tao and Xie, Lihua and Hong, Yiguang and Chai, Tianyou and Johansson, Karl H},
  journal={arXiv preprint arXiv:2306.00149},
  year={2023}
}

@article{hazan2007adaptive,
  title={Adaptive online gradient descent},
  author={Hazan, Elad and Rakhlin, Alexander and Bartlett, Peter},
  journal={Advances in neural information processing systems},
  volume={20},
  year={2007}
}

@InProceedings{pmlr-v37-daniely15,
  title = 	 {Strongly Adaptive Online Learning},
  author = 	 {Daniely, Amit and Gonen, Alon and Shalev-Shwartz, Shai},
  booktitle = 	 {Proceedings of the 32nd International Conference on Machine Learning},
  pages = 	 {1405--1411},
  year = 	 {2015},
  editor = 	 {Bach, Francis and Blei, David},
  volume = 	 {37},
  series = 	 {Proceedings of Machine Learning Research},
  address = 	 {Lille, France},
  month = 	 {07--09 Jul},
  publisher =    {PMLR}
}

@article{garber2021efficient,
  author  = {Dan Garber},
  title   = {Efficient online linear optimization with approximation algorithms},
  journal = {Mathematics of Operations Research},
  volume  = {46},
  number  = {1},
  pages   = {204--220},
  year    = {2021}
}

@inproceedings{kretzu2021revisiting,
  author    = {Ben Kretzu and Dan Garber},
  title     = {Revisiting Projection-free Online Learning: the Strongly Convex Case},
  booktitle = {Proceedings of The 24th International Conference on Artificial Intelligence and Statistics},
  series    = {Proceedings of Machine Learning Research},
  volume    = {130},
  pages     = {3592--3600},
  year      = {2021},
  editor    = {Arindam Banerjee and Kenji Fukumizu},
  publisher = {PMLR}
}

@inproceedings{jaggi2013revisiting,
  author    = {Martin Jaggi},
  title     = {Revisiting Frank-Wolfe: Projection-free sparse convex optimization},
  booktitle = {Proceedings of the 30th International Conference on Machine Learning (ICML)},
  volume    = {28},
  pages     = {427--435},
  year      = {2013}
}

@article{orabona2019modern,
  title={A modern introduction to online learning},
  author={Orabona, Francesco},
  journal={arXiv preprint arXiv:1912.13213},
  year={2019}
}

@inproceedings{wan2021projection,
  title={Projection-free online learning over strongly convex sets},
  author={Wan, Yuanyu and Zhang, Lijun},
  booktitle={Proceedings of the AAAI Conference on Artificial Intelligence},
  volume={35},
  pages={10076--10084},
  year={2021}
}

@inproceedings{mhammedi2022efficient,
  title={Efficient projection-free online convex optimization with membership oracle},
  author={Mhammedi, Zakaria},
  booktitle={Conference on Learning Theory},
  pages={5314--5390},
  year={2022},
  organization={PMLR}
}

@inproceedings{sarkar2025projection,
  title={Projection-free Algorithms for Online Convex Optimization with Adversarial Constraints},
  author={Sarkar, Dhruv and Chakrabartty, Aprameyo and Supantha, Subhamon and Dey, Palash and Sinha, Abhishek},
  booktitle={Proceedings of the 29th International Conference on Artificial Intelligence and Statistics},
  year={2026},
  url={https://arxiv.org/abs/2501.16919}
}

@inproceedings{mhammedi2025online,
  title={Online Convex Optimization with a Separation Oracle},
  author={Mhammedi, Zakaria},
  booktitle={Proceedings of Thirty Eighth Conference on Learning Theory},
  pages={4033--4077},
  year={2025},
  volume={291},
  publisher={PMLR}
}

@inproceedings{wang2025revisiting,
  title={Revisiting projection-free online learning with time-varying constraints},
  author={Wang, Yibo and Wan, Yuanyu and Zhang, Lijun},
  booktitle={Proceedings of the AAAI Conference on Artificial Intelligence},
  volume={39},
  pages={21339--21347},
  year={2025}
}

@inproceedings{Miyato2018spectral,
  author    = {Miyato, Takeru and Kataoka, Toshiki and Koyama, Masanori and Yoshida, Yuichi},
  title     = {Spectral Normalization for Generative Adversarial Networks},
  booktitle = {International Conference on Learning Representations},
  year      = {2018},
  url       = {https://openreview.net/forum?id=B1QRgziT-}
}

@book{zhou1996robust,
author = {Zhou, Kemin and Doyle, John C. and Glover, Keith},
title = {Robust and optimal control},
year = {1996},
isbn = {0134565673},
publisher = {Prentice-Hall, Inc.},
address = {USA}
}

@article{guo2022online,
  title={Online convex optimization with hard constraints: Towards the best of two worlds and beyond},
  author={Guo, Hengquan and Liu, Xin and Wei, Honghao and Ying, Lei},
  journal={Advances in Neural Information Processing Systems},
  volume={35},
  pages={36426--36439},
  year={2022}
}

@article{lu2025decentralized,
title={Decentralized Projection-free Online Upper-Linearizable Optimization with Applications to {DR}-Submodular Optimization},
author={Yiyang Lu and Mohammad Pedramfar and Vaneet Aggarwal},
journal={Transactions on Machine Learning Research},
issn={2835-8856},
year={2025}
}

@inproceedings{lu2026upper,
  title={Upper-Linearizability of Online Non-Monotone DR-Submodular Maximization over Down-Closed Convex Sets},
  author={Lu, Yiyang and Jadav, Haresh and Pedramfar, Mohammad and Singh, Ranveer and Aggarwal, Vaneet},
  booktitle={Proceedings of the 43rd International Conference on Machine Learning},
  year={2026}
}

@article{scarvelis2024nuclear,
  title={Nuclear norm regularization for deep learning},
  author={Scarvelis, Christopher and Solomon, Justin},
  journal={Advances in Neural Information Processing Systems},
  volume={37},
  pages={116223--116253},
  year={2024}
}
\bibliographystyle{tmlr}

\newpage
\appendix

\section{Useful lemmas}

Here we introduce some technical lemmas that are used in our proofs.

\begin{lemma}[Lemma~4.13 in~\citet{orabona2019modern}]
\label{lem:f}
Let $a_0 \geq 0$ be a real number, $N \geq 1$ be an integer, and $(a_t)_{t = 1}^N$ be a sequence of non-negative real numbers.
For any non-increasing function $f : [0, \infty) \to [0, \infty)$, we have
\begin{align*}
\sum_{t = 1}^N a_t f\left( a_0 + \sum_{i=1}^t a_i \right)
\leq \int_{a_0}^{a_0+\sum_{t = 1}^N a_t} f(x) dx.
\end{align*}
\end{lemma}

\begin{lemma}
\label{lem:scalar-stronvex}
    Assume $f$ is $\theta$-strongly convex and $g$ is convex. Let $a$ and $b$ be some non-negative real number. Then $a f + b g$ is $a\theta$-strongly convex.
\end{lemma}
\begin{proof}
    Given $ f $ is $ \theta $-strongly convex, its Hessian satisfies $\nabla^2 f(x) \succeq \theta I$, where $I$ stands for the identity matrix of appropriate dimensions.
    Since $g$ is convex, its Hessian satisfies $\nabla^2 g(x) \succeq 0$.
    Let $h(x) = af(x) + bg(x)$. 
    Then, its Hessian is given by $\nabla^2 h(x) = a \nabla^2 f(x) + b \nabla^2 g(x) \succeq a\theta I$, where $I$ stands for the identity matrix of appropriate dimensions.
    Thus, \( h(x) \) is strongly convex with parameter \( a\theta \).
\end{proof}


\section{Extended Preliminaries on Oracle Types}
\label{apx:oracles}

In classical projection-based algorithms, \textit{Projection Oracle} (PO/$\mathcal{O}_{P}$) and \textit{Convex Optimization Oracle} (COO/$\mathcal{O}_{CO}$) are standard feasibility mechanisms for actions chosen by the algorithm.
Given a closed convex set $\mathcal{K} \subseteq \mathbb{R}^n$ and a query point $y \in \mathbb{R}^n$, a PO returns the unique point $\x \in \mathcal{K}$ that minimizes the Euclidean distance to $\y$, i.e., $\mathcal{O}_{P}(\y, \mathcal{K}) = \arg\min_{\x \in \mathcal{K}} \| \y - \x \|_2^2$.
Given a general convex objective function $h: \mathbb{R}^n \rightarrow \mathbb{R}$ over $\mathcal{K}$,
a COO returns a global minimizer, i.e., $\mathcal{O}_{CO}(h, \mathcal{K}) \in \arg\min_{\x \in \mathcal{K}} \ h(\x)$.
In the context of COCO, COOs are often invoked to handle time-varying soft constraints $g_t(\x) \le 0$ without performing a direct projection. 
If $h(\x)$ is quadratic, a COO acts as a PO. 
If $h(\x)$ includes complex non-linear constraints, a COO represents a more computationally intensive subroutine than a simple projection.

The \textit{Linear Optimization Oracle} (LOO/$\mathcal{O}_{LO}$) and \textit{Separation Oracle} (SO/$\mathcal{O}_{S}$) are established approaches to move from projection-based algorithm to projection-free paradigms. 
The LOO exploits the fact that for many complex geometries, minimizing a linear function is vastly cheaper than minimizing a quadratic function (PO) or a general convex function (COO). Given a convex set $\mathcal{K}\subseteq \mathbb{R}^n$ and a linear objective vector $\BF{c} \in \mathbb{R}^n$ (typically a gradient or a direction vector), a LOO returns a vertex or point in $\mathcal{K}$ that minimizes the linear function: $\mathcal{O}_{LO}(\BF{c}, \mathcal{K}) \in \arg\min_{\x \in \mathcal{K}} \ \langle \BF{c}, \x \rangle $.
The SO is a more recent development, and another natural projection-free setting to access the action set. Given a convex set $\mathcal{K} \subseteq \mathbb{R}^n$ and a query point $\y \in \mathbb{R}^n$, it either asserts that $\y \in \mathcal{K}$, or it returns a separating hyperplane vector $\BF{g} \in \mathbb{R}^n$ such that for all $\x \in \mathcal{K}$, $\langle \BF{g}, \y - \x \rangle > 0$.
Although LOOs are more prevalent in the literature, it is worth noting the complementing nature between the two oracles in efficiency. \citet{garber2022new} gave the example of two unit balls: a nuclear norm ball $\mathcal{B}_* = \{ \mathbf{x} \in \mathbb{R}^{m \times n} \mid \|\mathbf{x}\|_* \leq 1 \}$, and a spectral norm ball $\mathcal{B}_2 = \{ \mathbf{x} \in \mathbb{R}^{m \times n} \mid \|\mathbf{x}\|_2 \leq 1 \}$. 
While the requirement of a full Singular Value Decomposition (SVD) makes Euclidean projection computationally prohibitive onto either set, the efficiency of projection-free oracles differs. The LOO is efficient for $\mathcal{B}_*$ (requiring only a rank-one SVD) but expensive for $\mathcal{B}_2$ where a full SVD is needed \citep{jaggi2013revisiting}. In contrast, the SO is expensive for $\mathcal{B}_*$ due to the subgradient for the nuclear norm (full SVD), but efficient for $\mathcal{B}_2$ since computing a subgradient for $g_2(\mathbf{X}) := \|\mathbf{X}\|_2 - 1$ requires only the top singular vector pair (rank-one SVD).


\section{ Infeasible Projection via a Separation Oracle}
\label{apx:ipso}

As mentioned in Section~\ref{sec:relatedWork}, our work builds upon the concept of infeasible projection, implemented via a separation oracle, pioneered by \citet{garber2022new} for the unconstrained OCO setting. We specifically adopt the variant from \citet{pedramfar2024linearizable} due to its favorable property of ensuring iterates remain within the constraint set $\K$, which is crucial for our analysis. Here we formally introduce the details of IP-SO algorithm in Algorithm~\ref{alg:ipso} and an important Lemma~\ref{lem:ipso}.

As mentioned in Section~\ref{sec:relatedWork} and Section~\ref{sec:prelim}, our work builds upon the concept of infeasible projection, implemented via a separation oracle, pioneered by \citet{garber2022new} for the unconstrained OCO setting. 

Formally, we say $\mathbf{y}' \in \mathbb{R}^n$ is an \emph{infeasible projection} of $\mathbf{y}$ onto $\mathcal{K}$ if for all $\mathbf{x} \in \mathcal{K}$, we have $\|\mathbf{y}'-\mathbf{x}\| \leq \|\mathbf{y}-\mathbf{x}\|$. Note that, unlike exact Euclidean projection, $\mathbf{y}'$ is not strictly assumed to be in $\mathcal{K}$ (though our chosen variant ensures it is). An algorithm returns an infeasible projection for set $\mathcal{K}$ if it takes an input point $\mathbf{y}$ and returns a $\mathbf{y}'$ that satisfies this property.

An infeasible projection can be implemented efficiently using a separation oracle. For a convex set $\mathcal{K} \subseteq \mathbb{R}^n$, a Separation Oracle ($\operatorname{SO}_{\mathcal{K}}$) is an oracle that, given a point $\mathbf{y} \in \mathbb{R}^n$, either confirms that $\mathbf{y} \in \mathcal{K}$ or returns a separating hyperplane vector $\mathbf{g}$ such that for all $\mathbf{x} \in \mathcal{K}$, $\langle \mathbf{y} - \mathbf{x}, \mathbf{g} \rangle > 0$.

Here, we specifically adopt the variant of the IP-SO algorithm described in \citet{pedramfar2024linearizable} due to its favorable property of ensuring iterates remain within the constraint set $\mathcal{K}$, which is crucial for our analysis. We include it as Algorithm~\ref{alg:ipso} for completeness. We use $\mathcal{P}_{\mathcal{K}, \mathbf{c}, r, \delta}$ (or simply $\mathcal{P}$ when there is no ambiguity) to denote this algorithm. This variant slightly differs from the original version described in \citet{garber2022new} in that it does not require the assumption that $\mathcal{K}$ contains a $d$-dimensional ball centered at the origin, nor does it require $\mathbf{0} \in \mathcal{K}$, and has been adopted in several works \cite{lu2025decentralized,lu2026upper}.

The key idea of the algorithm is to provide an infeasible projection onto $\mathcal{K}_\delta$ that is contained within $\mathcal{K}$. This guarantees that the infeasible projection is not too far from the exact projection, since any point in $\mathcal{K}$ is not too far from $\mathcal{K}_\delta$. Lemma~\ref{lem:ipso} ensures that the algorithm output is an infeasible projection within a bounded number of iterations (or queries of the separation oracle).

\begin{algorithm}[ht] 
\caption{Infeasible Projection via a Separation Oracle (IP–SO) $\mathcal{P}_{\mathcal{K}, \mathbf{c}, r, \delta}$ \cite{pedramfar2024linearizable}}
\label{alg:ipso}
{\small
\begin{algorithmic}[1]
  \Require Constraint set $\C{K}$, $\BF{c} \in \op{relint}(\K)$, $r = r_{\K, \BF{c}}$, shrinking parameter $\delta \in [0,1)$, initial point $\y_0$
  \State $\mathbf{y}_1 \gets \mathbf{P}_{\operatorname{aff}(\mathcal{K})}(\mathbf{y}_0)$ \Comment{$\y_1$ is projection of $\y_0$ over $\BB{B}_{D}(\BF{c}) \cap \op{aff}(\C{K})$}
  \State  $\y_2 \gets \BF{c} + \frac{\y_1 - \BF{c}}{\max\{1, \|\y_1\|/D \}}$  
  \For{$i=1,2,\dots$}
    \State Call $\operatorname{SO}_{\mathcal{K}}$ with input $\mathbf{y}_i$
    \If{$\mathbf{y}_i \notin \mathcal{K}$}
      \State Set $\g_i$ to be the hyperplane returned by $\op{SO}_{\C{K}}$ \Comment{$\forall \x \in \C{K}$, $\langle \y_i - \x, \g_i \rangle > 0$}
      \State $\mathbf{g}'_i \gets \mathbf{P}_{\operatorname{aff}(\mathcal{K})-\mathbf{c}}(\mathbf{g}_i)$
      \State $\mathbf{y}_{i+1} \gets \mathbf{y}_i - \delta r\, \dfrac{\mathbf{g}'_i}{\|\mathbf{g}'_i\|}$
    \Else
      \State \Return $\mathbf{y}\gets \mathbf{y}_i$
    \EndIf
  \EndFor
\end{algorithmic}
}
\end{algorithm}

\begin{lemma}[Lemma 5 in~\citet{pedramfar2024linearizable}]
\label{lem:ipso}
Given a constrained set $\K$, a shrinking parameter $\delta \in [0, 1)$ (where $r$ is as previously described), and an initial point $\y_0$, Algorithm~\ref{alg:ipso} stops after at most $\frac{\op{dist}(\y_0,\K_\delta)^2-\op{dist}(\y,\K_\delta)^2}{\delta^2r^2}+1$ iterations and returns $\y \in \K$ such that for all $\x \in \K_\delta$, we have $\Vert \y-\x \Vert \leq \Vert \y_0 - \x \Vert$.
\end{lemma}


\section{Proof of Theorem~\ref{thm:unconstrained-regret}}
\label{apx:unconstrained-regret}

\begin{proof}
Let $\xs_0\in \op{argmin}_{\x\in\K}\sum_{t=1}^T f_t(\x)$ and $\txs_0=(1-\delta)\xs_0+\delta{\BF{c}} \in \K_\delta$.
Thus, the regret of the proposed algorithm is given as
\begin{align}
\label{eq:decomposeRegret}
\mathtt{Regret}(T) 
&= \sumT \left( f_t(\x_t) - f_t(\xs_0) \right) \nonumber \\
& = \sumT \left( f_t(\x_t) - f_t(\txs+0) \right) \nonumber \\
& + \sumT \left( f_t(\txs_0) - f_t(\xs_0) \right).
\end{align}
We bound the two terms separately.

First we consider the term $\sum_{t=1}^T \left( f_t(\x_t) - f_t(\txs_0) \right)$.
From Lemma~\ref{lem:ipso}, we have for $\txs_0 \in \K_\delta$,
\begin{align*}
\Vert \x_{m+1}-\txs_0 \Vert ^2 
& \leq \Vert \x_m - \eta_m \bn_m -\txs_0 \Vert ^2 \\
&\leq \Vert \x_m - \txs_0 \Vert ^2 + \eta_m^2 \Vert \bn_m \Vert^2 \\
& - 2\eta_m \langle \bn_m, \x_m - \txs_0 \rangle.
\end{align*}
Rearranging, we have
\begin{align}
\langle \bn_m, \x_m - \txs_0 \rangle 
&\leq \frac{1}{2\eta_m} (\Vert \x_m - \txs_0 \Vert ^2- \Vert \x_{m+1}-\txs_0 \Vert ^2) \nonumber \\
&+ \frac{\eta_m}{2} \Vert \bn_m \Vert^2
\end{align}
Summing this over $m$ from $1$ to $T/K$, we have
\begin{align*}
&\sum\limits_{m=1}^{T/K} \langle \bn_m , \x_m - \txs_0 \rangle  
\leq \frac{\Vert \x_1 - \txs_0 \Vert ^2}{2\eta_1} \\
& \quad  + \sum\limits_{m=2}^{T/K} (\frac{1}{2\eta_m}-\frac{1}{2\eta_{m-1}}) \Vert \x_m - \txs_0 \Vert ^2 
+ \sum\limits_{m=1}^{T/K} \frac{\eta_m}{2} \Vert \bn_m \Vert^2 \\
&\quad \leq \frac{D^2}{2\eta_1} 
+ D^2 \sum\limits_{m=2}^{T/K} \left| \frac{1}{2\eta_m}-\frac{1}{2\eta_{m-1}} \right| 
+ \sum\limits_{m=1}^{T/K} \frac{\eta_m}{2} \Vert \bn_m \Vert^2 \\
&\quad = \frac{D^2}{2\eta_1} 
+ D^2 \sum\limits_{m=2}^{T/K} \left( \frac{1}{2\eta_m}-\frac{1}{2\eta_{m-1}} \right) 
+ \sum\limits_{m=1}^{T/K} \frac{\eta_m}{2} \Vert \bn_m \Vert^2 \\
&\quad = \frac{D^2}{2\eta_{T/K}}
+ \sum\limits_{m=1}^{T/K} \frac{\eta_m}{2} \Vert \bn_m \Vert^2,
 \end{align*}
where we used the facts that $\Vert \x_m - \txs_0 \Vert \leq D$ and that $\eta_m$ is non-increasing.
By the convexity of $f_t$, we have $f_t (\x_t) - f_t(\txs_0) \leq \langle \nabla_t, \x_t - \z \rangle$. Summing this over $t$, we have
\begin{align}
\label{eq:xt2txs}
&\sumT f_t (\x_t) - \sumT f_t (\txs_0) 
\leq \sumT \langle \nabla_t, \x_t - \txs_0 \rangle \nonumber \\
&\quad \quad = \sum\limits_{m=1}^{T/K} \sum\limits_{t\in\C{T}_m} \langle \nabla_t, \x_t - \txs_0 \rangle \nonumber\\
&\quad \quad= \sum\limits_{m=1}^{T/K} K \langle \bn_m, \x_m - \txs_0 \rangle \nonumber \\
&\quad \quad \leq  \frac{K D^2}{2\eta_{T/K}} + \frac{K}{2} \sum\limits_{m=1}^{T/K} \eta_m \normdm^2.
\end{align}
Note that if $\eta_t$ is chosen to be a constant, $\eta_m = \frac{D}{\sqrt{\sum_{m=1}^{T/K} \Vert \bn_m \Vert^2}}$ minimizes the right hand side. However, this step size could not be chosen since at a given $m$, the future $\bn_m$'s are not known. 
Thus, we use adaptive step size in this work, where $\eta_m = \frac{D}{\sqrt{\epsilon + \sum^m_{\tau=1}\Vert \bn_\tau \Vert^2}}$.
Using this, Equation~\eqref{eq:xt2txs} is further bounded as 
\begin{align*}
& \sumT f_t (\x_t) - \sumT f_t (\txs_0) 
\leq \frac{K D^2}{2 \eta_{T/K}} + \frac{K}{2} \sum\limits_{m=1}^{T/K} \eta_m \normdm^2 \nonumber \\
& \leq \frac{K D}{2}\sqrt{\epsilon+\sum\limits_{m=1}^{T/K}\Vert \bn_m \Vert^2} 
    + \frac{K D}{2} \sum\limits_{m=1}^{T/K} \frac{ \normdm^2}{\sqrt{\epsilon + \sum^m_{\tau=1}\Vert \bn_\tau \Vert^2}}. 
\end{align*}
By applying Lemma~\ref{lem:f} to $f(x) = 1/\sqrt{x}$, $a_0 = \epsilon$, $N = T/K$, and $a_m = \Vert \bn_m \Vert^2$ for $m\ge 1$, we see that 
\begin{align*}
&\sum_{m=1}^{T/K} \frac{\normdm^2}{\sqrt{\epsilon + \sum^m_{\tau=1}\Vert \bn_\tau \Vert^2}} 
= \sum_{m=1}^{T/K} a_m f\left( a_0 + \sum^m_{\tau=1} a_\tau \right) \\
&\qquad \leq \int_{a_0}^{a_0+\sum_{t = \tau}^{T/K} a_\tau} f(x) dx 
= 2 \sqrt{\epsilon+\sum_{m=1}^{T/K} \|\bn_m\|^2} - 2 \sqrt{\epsilon}..
\end{align*}
Thus
\begin{align}
\label{eq:xt2txs1}
   \sum_{t=1}^T f_t(\x_t) - \sum_{t=1}^T f_t(\tilde{\mathbf{x}}_0^*) &\le \frac{KD}{2} \sqrt{\epsilon + \sum_{m=1}^{T/K} \|\bar{\nabla}_m\|^2} + \frac{KD}{2} \left( 2\sqrt{\epsilon + \sum_{m=1}^{T/K} \|\bar{\nabla}_m\|^2} - 2\sqrt{\epsilon} \right) \nonumber \\
    &\le \frac{3KD}{2} \sqrt{\epsilon + \sum_{m=1}^{T/K} \|\bar{\nabla}_m\|^2} - KD\sqrt{\epsilon} \nonumber \\
    &\le \frac{3KD}{2} \sqrt{\sum_{m=1}^{T/K} \|\bar{\nabla}_m\|^2} + \frac{3KD}{2} \sqrt{\epsilon} - KD\sqrt{\epsilon} \nonumber \\
    &= \frac{3KD}{2} \sqrt{\sum_{m=1}^{T/K} \|\bar{\nabla}_m\|^2} + \frac{KD}{2} \sqrt{\epsilon}.
\end{align}

On the other hand, the expression $\sumT\left(f_t(\txs_0)-f_t(\xs_0)\right)$ may be bounded as
\begin{align}
\label{eq:txs2xs}
\sumT\left(f_t(\txs_0)-f_t(\xs_0)\right)
&\overset{(a)}{\leq} \sumT G_t \Vert \txs_0-\xs_0 \Vert \nonumber\\
& \overset{(b)}{\leq} \delta D \sumT G_t,
\end{align}
where (a) is because $f_t$ are $G_t$-Lipschitz continuous, and (b) is due to the fact that 
\begin{align*}
\Vert \txs_0-\xs_0 \Vert= \Vert (1-\delta)\xs_0+\delta\BF{c}-\xs_0 \Vert = \delta\Vert  \xs_0 - \BF{c} \Vert \leq \delta D.
\end{align*}
The claim now follows by using the bounds in \eqref{eq:xt2txs1} and \eqref{eq:txs2xs} together with \eqref{eq:decomposeRegret}.

Next we examine the number of calls to the separation oracles through the infeasible projection algorithm.
Using triangle inequality, $\forall \y, \y_0 \in \K$, we have
\begin{align*}
\op{dist}(\y_0,\K_\delta)
&= \min_{\z \in \K_\delta} \| \y_0 - \z \| \\
&\leq \min_{\z \in \K_\delta} \left( \| \y - \z \| + \| \y  - \y_0 \| \right) \\
&= \op{dist}(\y,\K_\delta) + \| \y  - \y_0 \|.
\end{align*}
Therefore
\begin{align*}
&\op{dist}(\y_0,\K_\delta)^2 - \op{dist}(\y,\K_\delta)^2 \\
&\leq \left( \op{dist}(\y,\K_\delta) + \| \y  - \y_0 \| \right)^2 - \op{dist}(\y,\K_\delta)^2 \\
&= \| \y  - \y_0 \|^2 + 2 \| \y  - \y_0 \| \op{dist}(\y,\K_\delta) \\
&\leq \| \y  - \y_0 \|^2 + 2 D \delta \| \y  - \y_0 \|,
\end{align*}
where the last inequality follows from that fact that, for all $\z \in \K$, we have $\tilde{\z} \triangleq (1-\delta)\z+\delta{\BF{c}} \in \K_\delta$ and therefore
\begin{align*}
\op{dist}(\z,\K_\delta)
&\leq \| \z - \tilde{\z} \|
= \| \z - (1-\delta)\z-\delta{\BF{c}} \| \\
&= \delta \| \z- \BF{c} \|
\leq D \delta.
\end{align*}
Thus, using Lemma~\ref{lem:ipso}, we see that the number of calls in Algorithm~\ref{alg:general} is bounded by
\begin{align}
\label{eq:oracle-calls}
&\sum_{m = 1}^{T/K - 1}\left( \frac{\op{dist}(\x_m - \eta_m \bn_m,\K_\delta)^2-\op{dist}(\x_{m+1},\K_\delta)^2 }{\delta^2r^2}+1 \right) \nonumber \\
&= \frac{1}{\delta^2r^2} \sum_{m = 1}^{T/K - 1}\left( \op{dist}(\x_m - \eta_m \bn_m,\K_\delta)^2-\op{dist}(\x_{m+1},\K_\delta)^2 \right) \nonumber \\
&\qquad+ \frac{T}{K} - 1 \nonumber \\
&\leq \frac{1}{\delta^2r^2} \sum_{m = 1}^{T/K - 1} ( \eta_m^2 \| \bn_m \|^2 + 2D\delta \eta_m \| \bn_m \| \nonumber \\
&\qquad+ \op{dist}(\x_m, \K_\delta)^2-\op{dist}(\x_{m+1},\K_\delta)^2 )
+ \frac{T}{K} \nonumber \\
&= \frac{1}{\delta^2r^2} \sum_{m = 1}^{T/K - 1} \left( \eta_m^2 \| \bn_m \|^2 + 2D\delta \eta_m \| \bn_m \| \right) \nonumber \\
&\qquad + \frac{1}{\delta^2r^2} \left( \op{dist}(\x_1,\K_\delta)^2-\op{dist}(\x_{T/K},\K_\delta)^2 \right)
+ \frac{T}{K} \nonumber \\
&\leq \frac{1}{\delta^2r^2} \sum_{m = 1}^{T/K} \left( \eta_m^2 \| \bn_m \|^2 + 2D\delta \eta_m \| \bn_m \| \right) \nonumber \\
& \qquad + {\color{blue}}\frac{D^2}{r^2} 
+ \frac{T}{K},
\end{align}
where we used the fact that $\op{dist}(\x_1,\K_\delta) \leq D\delta$ and that $\left( \eta_m^2 \| \nabla_m \|^2 + 2D\delta \eta_m \| \nabla_m \| \right)$ is positive when $m=\frac{T}{K}$ in the last inequality.

If $\epsilon > 0$, then using Lemma~\ref{lem:f} for $f(x) = 1/x$, $N = T/K$, $a_0 = \epsilon$, and $a_i = \| \bn_i \|^2$, we see that
\begin{align*}
\sum_{m = 1}^{T/K} \frac{\| \bn_m \|^2}{\epsilon + \sum_{\tau=1}^{m}\| \bn_m \|^2} 
&= \sum_{m = 1}^{T/K} a_m f\left( a_0 + \sum_{\tau=1}^{m} a_\tau \right) \\
&\leq \int_{a_0}^{a_0+ \sum_{m = 1}^{T/K} a_m} f(x) dx \\
&= \log\left( \epsilon + \sum_{m = 1}^{T/K} \| \bn_m \|^2 \right) - \log(\epsilon) \\
&= \log\left( 1 + \frac{1}{\epsilon} \sum_{m = 1}^{T/K} \| \bn_m \|^2 \right).
\end{align*}
Therefore
\begin{align*}
\sum_{m = 1}^{T/K} \eta_m^2 \| \bn_m \|^2 
&= D^2 \sum_{m = 1}^{T/K} \frac{\| \bn_m \|^2}{\epsilon + \sum_{\tau=1}^{m}\| \bn_m \|^2} \\
&\leq D^2\log\left( 1 + \frac{1}{\epsilon} \sum_{m = 1}^{T/K} \| \bn_m \|^2 \right) \\
&\leq 2D^2\log\left( 1 + \frac{M_1 T}{\epsilon K} \right),
\end{align*}
where we used the fact that $\bn_m=\frac{1}{K}\sum_{t\in\C{T}_m}\nabla_t$ is bounded by $M_1$.
Thus, using Cauchy–Schwarz inequality, we see that
\begin{align*}
\left( \sum_{m = 1}^{T/K} \eta_m \| \bn_m \| \right)^2
&\leq 
\left( \sum_{m = 1}^{T/K} 1 \right)
\left( \sum_{m = 1}^{T/K} \eta_m^2 \| \bn_m \|^2 \right) \\
&\leq \frac{2D^2T}{K} \log\left( 1 + \frac{M_1 T}{\epsilon K} \right).
\end{align*}
Therefore, the number of calls is bounded by
\begin{align*}
&\frac{1}{\delta^2r^2} \sum_{m = 1}^{T/K - 1} \left( \eta_m^2 \| \nabla_m \|^2 + 2D\delta \eta_m \| \nabla_m \| \right)
+ \frac{D^2}{r^2} 
+ \frac{T}{K} \\
&\leq \frac{2D^2}{\delta^2r^2} \log\left( 1 + \frac{M_1 T}{\epsilon K} \right)
+ \frac{2 \sqrt{2} D^2}{r^2 \delta} \sqrt{\frac{T}{K} \log\left( 1 + \frac{M_1 T}{\epsilon K} \right)} \\
& \qquad + \frac{D^2}{r^2} 
+ \frac{T}{K} \\
&= \C{O}( (D/r)^2\delta^{-2} \log(T) \\
& \qquad + (D/r)\delta^{-1} (\log(T))^{1/2} (T K^{-1})^{1/2} 
+ T K^{-1} ) \\
&= \C{O}\left( (D/r)^2\delta^{-2} \log(T) 
+ T K^{-1} \right),
\end{align*}
where we used the fact that $\epsilon$ is a constant independent of $T$ and $\sqrt{a b} = \C{O}(a + b)$ in the last equality.

Note that, if we only assume that $\epsilon \geq 0$, then the above regret bound becomes infinite as it contains the term $\log\left( 1 + \frac{M_1 T}{\epsilon K} \right)$.
However, we may instead bound $\sum_{m = 1}^{T/K} \frac{\| \bn_m \|^2}{\epsilon + \sum_{\tau=1}^{m}\| \bn_m \|^2}$ with $T/K$, since each term is bounded by $1$.
Thus, using Cauchy–Schwarz inequality, we see that
\begin{align*}
\left( \sum_{m = 1}^{T/K} \eta_m \| \bn_m \| \right)^2
&\leq 
\left( \sum_{m = 1}^{T/K} 1 \right)
\left( \sum_{m = 1}^{T/K} \eta_m^2 \| \bn_m \|^2 \right) \\
&\leq \frac{2D^2 T}{K} \cdot \frac{T}{K}.
\end{align*}
Therefore, if we allow $\epsilon=0$, i.e., if we only assume $\epsilon \geq 0$, the number of calls is bounded by
\begin{align*}
&\frac{1}{\delta^2r^2} \sum_{m = 1}^{T/K - 1} \left( \eta_m^2 \| \nabla_m \|^2 + 2D \delta \eta_m \| \nabla_m \| \right)
+ \frac{D^2}{r^2} 
+ \frac{T}{K} \\
&\quad\leq \frac{2D^2}{\delta^2r^2} \frac{T}{K}
+ \frac{2 \sqrt{2} D^2}{r^2 \delta} \frac{T}{K}
+ \frac{D^2}{r^2} 
+ \frac{T}{K} \\
&\quad= \C{O}\left( (D/r)^2\delta^{-2} T K^{-1} 
+ (D/r)\delta^{-1} T K^{-1}
+ T K^{-1} \right) \\
&\quad= \C{O}\left((D/r)^2 T K^{-1} \delta^{-2} \right).
\qedhere
\end{align*}
\end{proof}


\section{Analysis of Base Algorithm for Strongly convex cost}
\label{apx:unconstrained-strongly}

\begin{theorem}
\label{thm:unconstrained-strongly}
Assume functions $f_t$ are $G_t$-Lipschitz continuous and $\theta$-strongly convex for some constant $\theta > 0$. Then Algorithm~\ref{alg:general} ensures that the regret is bounded as
\begin{align*}
\mathtt{Regret}(T) \leq \frac{1}{2 \theta} \sum\limits_{m=1}^{T/K} \sum\limits_{t\in\C{T}_m} \frac{\normdt^2}{m} + \frac{\delta }{r} D \sumT G_t,
\end{align*}
where $\delta$, $K$, and $D$ are defined as in Theorem~\ref{thm:unconstrained-regret}.
Moreover, if we assume all $G_t \leq M_1$ for some constant $M_1 \geq 0$, then
this requires $\C{O}\left( \delta^{-2} + \delta^{-1}\log(T) + T K^{-1} \right)$ calls to the separation oracle.
\end{theorem}

\begin{proof}
Let $\xs_0\in \op{argmin}_{\x\in\K}\sum_{t=1}^T f_t(\x)$ and $\txs_0=(1-\frac{\delta}{r})\xs_0+\frac{\delta}{r}{\BF{c}} \in \K_\delta$.
We may decompose the regret in the same manner as Equation~\eqref{eq:decomposeRegret} to see that
\begin{align*}
\mathtt{Regret}(T) &= \sumT \left( f_t(\x_t) - f_t(\xs_0) \right) \\
&= \sumT \left( f_t(\x_t) - f_t(\txs_0) \right) \\
&+ \sumT \left( f_t(\txs_0) - f_t(\xs_0) \right).
\end{align*}
Since we assume $f_t$ to be $G_t$-Lipschitz continuous, similar to Equation~\eqref{eq:txs2xs}, we have
\begin{align}
\label{eq:strconv-lip-cons}
    \sumT \left( f_t(\txs_0) - f_t(\xs_0) \right) \leq \frac{\delta D}{r} \sumT G_t.
\end{align}
The infeasible projection operator ensures that $\Vert \x_{t+1}-\txs_0 \Vert \leq \Vert \y_t - \txs_0 \Vert$, extending Theorem 4.1 by \cite{hazan2007adaptive}, we have
\begin{align*}
    \sumT \left( f_t(\x_t) - f_t(\txs_0) \right) 
    \leq \frac{1}{2} \sum\limits_{m=1}^{T/K} \sum\limits_{t\in\C{T}_m} \frac{G_t^2}{m \theta}
\end{align*}
Thus, we conclude that 
\begin{align*}
    \mathtt{Regret}(T) \leq \frac{1}{2 \theta} \sum\limits_{m=1}^{T/K} \sum\limits_{t\in\C{T}_m} \frac{G_t^2}{m} + \frac{\delta D}{r} \sumT G_t
\end{align*}
Note that if we assume all $G_t\leq M_1$, thus all functions are $M_1$-Lipschitz, then 
\begin{align*}
    \mathtt{Regret}(T) &\leq \frac{M_1^2 K}{2 \theta} \sum\limits_{m=1}^{T/K} \frac{1}{m} + \frac{\delta D M_1 T}{r}  \\
    &\leq \frac{M_1^2 K}{2 \theta} (1+\log(\frac{T}{K})) + \frac{\delta D M_1 T}{r}
\end{align*}
where the last is due to the sum of harmonic series. 
If $\delta = \Theta ( T^{-\beta}\op{log}(T))$, $K=\Theta(T^{1-\beta})$, then $\mathtt{Regret}(T)=\C{O}(T^{1-\beta}\log(T))$.

Next we examine the number of calls to the separation oracles. Similar to the convex case, the number of calls to the separation oracle is bounded by Equation~\eqref{eq:oracle-calls}:
\begin{align*}
&\sum_{m = 1}^{T/K - 1}\left( \frac{\op{dist}(\x_m - \eta_m \bn_m,\K_\delta)^2-\op{dist}(\x_{m+1},\K_\delta)^2 }{\delta^2}+1 \right) \nonumber \\
&\quad\leq \frac{1}{\delta^2} \sum_{m = 1}^{T/K} \left( \eta_m^2 \| \bn_m \|^2 + \frac{2 D \delta}{r} \eta_m \| \bn_m \| \right)
+ \frac{D}{r \delta} 
+ \frac{T}{K},
\end{align*}
but for the strongly convex functions, $\eta_m=\frac{1}{m\theta}$. Thus, if we assume all Lipschitz parameters are bounded by $M_1$, then we have
\begin{align*}
    \sum_{m = 1}^{T/K} \eta_m^2 \| \bn_m \|^2 \leq \frac{M_1^2}{\theta^2} \sum_{m = 1}^{T/K} \frac{1}{m^2} \leq \frac{2 M_1^2 }{\theta^2}
\end{align*}
where the last inequality follows from the fact that $\sum_{i = 1}^\infty i^{-2} < 2$.
On the other hand,
\begin{align*}
    \sum_{m = 1}^{T/K} \eta_m \| \bn_m \| 
    \leq \frac{M_1}{\theta} \sum_{m = 1}^{T/K} \frac{1}{m} 
    \leq \frac{M_1}{\theta}\left( \log\left( \frac{T}{K} \right) + 1 \right)
\end{align*}
where the last inequality follows from the fact that $\sum_{i = 1}^N i^{-1} < \log(N) + 1$.

Combining the results, we see that if all functions are $M_1$-Lipchitz, we have the number of calls to the separation oracles may be bounded as 
\begin{align*}
&\sum_{m = 1}^{T/K - 1}\left( \frac{\op{dist}(\x_m - \eta_m \bn_m,\K_\delta)^2-\op{dist}(\x_{m+1},\K_\delta)^2 }{\delta^2}+1 \right) \nonumber \\
&\quad\leq \frac{1}{\delta^2}  \left( \frac{2 M_1^2}{\theta^2} +  \frac{2 D \delta}{r} \frac{M_1}{\theta}\left( \log\left( \frac{T}{K} \right) + 1 \right) \right)
+ \frac{D}{r \delta} 
+ \frac{T}{K} \\
&\quad = \C{O}\left( \delta^{-2}
+ \delta^{-1}\log(T)
+ \delta^{-1}
+ T K^{-1} \right) \\
&\quad = \C{O}\left( \delta^{-2}
+ \delta^{-1}\log(T)
+ T K^{-1} \right). \qedhere
\end{align*}

\end{proof}

\begin{corollary}
\label{cor:strcon}
    In Theorem~\ref{thm:unconstrained-strongly}, if we assume all $G_t\leq M_1$ for some constant $M_1$, then we have
    \begin{align*}
    \mathtt{Regret}(T) 
    \leq  \frac{M_1^2 K}{2 \theta} (1+\log(\frac{T}{K})) + \frac{\delta D M_1 T}{r}.
    \end{align*}
    In particular, if we choose $\delta=\Theta(T^{-\beta}\op{log}(T))$ and $K=\Theta(T^{1-\beta})$ for some $\beta \in (0, 1]$, then $\mathtt{Regret}(T)= \C{O}(T^{1-\beta}\log(T))$
    with $\C{O}(T^{2\beta}(\log(T))^{-2})$ calls to the separation oracle.
\end{corollary}


\section{Proof of Theorem~\ref{thm:main}}
\label{apx:constrained-main}

The proof proceeds by connecting the performance of $\mathtt{BAGEL}$ on the COCO problem to the performance of its base OCO algorithm (Theorem 1) on the sequence of surrogate costs. The first step is bounding surrogate regret. We apply Theorem 1 directly to the sequence of surrogate functions $\{\hf_t\}$ to obtain a bound on the surrogate regret, $\mathtt{R}(\hf)$. This bound depends on the norms of the surrogate gradients. The second step is bounding surrogate gradients. We then bound the surrogate gradient norms in terms of the Lyapunov function's derivative, $\Phi'(Q_T)$, which effectively captures the worst-case cumulative violation. The third step is relating surrogate and true regret. We leverage the definition of $\hf_t$ and the convexity of $\Phi$ to establish the key inequality: $\mathtt{R}(\hf)\geq \mathtt{R}(\tf)+\Phi(Q_T)$. This inequality forms a bridge, linking the (boundable) surrogate regret to the two quantities we ultimately wish to control: the true regret $\mathtt{R}(\tf)$ and the total violation (captured by $\Phi(Q_T)$). Step 4: Parameter Tuning. By combining these inequalities and carefully choosing our hyperparameters ($\lambda,\gamma, \K, \delta$) to balance the resulting terms, we can solve for the final bounds on $\mathtt{Regret}(T)$ and $\mathtt{CCV}(T)$.

We start with bounding $\mathtt{R}(\hf)$:
\begin{lemma}
\label{lem:surrogate-regret}
Assume surrogate functions $\hf_t$ are convex and $G_t$-Lipschitz, then Algorithm~\ref{alg:constrained} ensures that the term $\mathtt{R}(\hf)$ defined as follows is bounded:
\begin{align*}
    \mathtt{R}(\hf) &\triangleq \sum_{t=1}^T \hf_t(\x_t) - \sum_{t=1}^T \hf_t(\x^*) \\
    &\leq \frac{3}{2}DK \sqrt{\sum_{m=1}^{T/K} \Vert \bn_m \Vert^2} + D \delta \sumT G_t + \frac{KD}{2}\sqrt{\epsilon},
\end{align*}
with $\tilde{\C{O}}(T^{2\beta})$ total calls to the SO, where $\x^*=\arg\min_{\x\in\K}\sum_{t=1}^T f_t(\x)$, $\bn_m$ is defined in Algorithm~\ref{alg:general}(line~9), $D$ is the diameter of the action set $\K$, $K$ is the block size, $\delta \in [0,1)$ is the shrinking parameter.
\end{lemma}
\begin{proof}
Let $\x^* = \arg\min_{\x \in \mathcal{K}^*} \sum_{t=1}^T f_t(\x)$ be the minimizer of the original costs over the strictly feasible set, and $\mathbf{x}_0^* = \arg\min_{\x \in \mathcal{K}} \sum_{t=1}^T \hat{f}_t(\x)$ be the minimizer of the surrogate functions. By definition of the minimum, for any $x \in \mathcal{K}$, including the strictly feasible original-cost minimizer $\x^*$, we have $\sum_{t=1}^T \hat{f}_t(\mathbf{x}_0^*) \le \sum_{t=1}^T \hat{f}_t(\x^*)$. Adding the algorithm's cumulative surrogate cost to both sides yields:
\begin{align*}
    \sum_{t=1}^T \hat{f}_t(\x_t) - \sum_{t=1}^T \hat{f}_t(\x^*) \le \sum_{t=1}^T \hat{f}_t(\x_t) - \sum_{t=1}^T \hat{f}_t(\mathbf{x}_0^*)
\end{align*}
Because Theorem~\ref{thm:unconstrained-regret} bounds the regret against the sequence minimizer $\mathbf{x}_0^*$, it mathematically upper bounds the right side of this inequality. Therefore, the exact same bound applies to the regret against $\x^*$ on the left side. 
\end{proof}

Next, we aim to bound $\mathtt{R}(\tf)$, using the relationship between $\hft$ and $\tft$, and replacing the $\Vert \bn_m \Vert ^2$ term in Lemma~\ref{lem:surrogate-regret}. The results are provided below in Lemma~\ref{lem:sur2real} with proof.
\begin{lemma}
\label{lem:sur2real}
    Assume $\Phi(\cdot):\BB{R}^+\mapsto\BB{R}^+$ is a non-decreasing, differentiable, and convex potential function satisfies $\Phi(0)=0$. Assume cost functions $f_t$ and constraint function $g_t$ are $M_1$-Lipschitz.
    Define $\mathtt{R}(\tf) \triangleq \sum_{t=1}^T \tf_t(\x_t) - \sum_{t=1}^T \tf_t(\x^*)$ where $\x^* = \arg\min_{x \in \mathcal{K}^*} \sum_{t=1}^T f_t(\x)$ and Algorithm~\ref{alg:constrained} ensures that
    \begin{equation}
    \label{eq:sur2real:1}
        \Phi (Q_T) + \mathtt{R}(\tf) \leq \mathtt{R} (\hf)
    \end{equation}
    and
    \begin{equation}
    \label{eq:sur2real:2}
        \mathtt{R}(\hat{f}) \leq \gamma M_1 D (1+ \Phi'(Q_T)) (\delta T + \frac{3}{\sqrt{2}} \sqrt{TK}) + \frac{KD}{2}\sqrt{\epsilon}
    \end{equation}
    where $\delta \in [0,1)$ is the shrinking parameter, $K$ is the block size, $D$ is the diameter of $\K$, and the definition of $\bn_m$ is given in Algorithm~\ref{alg:general} (line 9), $\gamma$ is the processing parameter.
\end{lemma}
\begin{proof}
    Note that $\hft(\xt)=\tft(\xt)+\Phi'(\qt)\tgt(\x_t)$, where $\tft=\gamma f_t$, $\tgt=\gamma (g_t)^+$, and $\Phi(\cdot)$ is a non-decreasing, differentiable, convex function satisfying $\Phi(0)=0$. By convexity of $\Phi(\cdot)$, we have
    \begin{align*}
    \Phi(\qt) 
    &\leq \Phi(\qtt) + \Phi'(\qt)(\qt-\qtt) \\
    &= \Phi(\qtt) + \Phi'(\qt)\tgt(\xt).
    \end{align*}
    Rearranging, we have $\Phi(Q_t) - \Phi(Q_{t-1}) \le \Phi'(Q_t)\tilde{g}_t(\x_t)$.
    Because $\x^*$ is defined as the optimal action within the strictly feasible set $\mathcal{K}^*$, it inherently satisfies $g_t(\x^*) \le 0$ for all $t$. Therefore, $\tilde{g}_t(\x^*) = \gamma(g_t(\x^*))^+ = 0$, which yields $\hat{f}_t(\x^*) = \tilde{f}_t(\x^*)$. Plugging this into the above inequality, we have
    \begin{align*}
        &\hft(\xt) - \hft(\xs) = \left(\tft(\xt)-\tft(\xs) \right) \\
        & \qquad+ \Phi'(\qt)(\tgt(\x_t)-\tgt(\x^*)) \\ 
        & \geq \left(\tft(\xt) - \tft(\xs) \right) + \Phi'(\qt)\tgt(\x_t) \\
        & \geq \left(\tft(\xt) - \tft(\xs) \right) + \Phi(\qt) - \Phi(\qtt).
    \end{align*}    
    Summing the above inequality over $t\in[T]$, with $Q_0=0$ and $\Phi(0)=0$, we have 
    \begin{align*}
        \mathtt{R}(\hf) \geq \mathtt{R}(\tf) + \Phi(Q_T),
    \end{align*}
    which gives Equation~\eqref{eq:sur2real:1} in Lemma~\ref{lem:sur2real}.
    
    Since both $f$ and $g$ are assumed to be $M_1$-Lipschitz continuous, we have 
    \begin{align}
    \label{eq:Gt}
         \normdt &= \Vert \nabla \hft \Vert 
    \leq \Vert \nabla \tft \Vert + \Phi'(\qt) \Vert \nabla \tgt \Vert \nonumber \\ 
    &\leq \gamma M_1(1+\Phi'(\qt)).
    \end{align}
    Thus, surrogate functions $\hft$ are $G_t$-Lipschitz continuous, where $G_t = \gamma M_1(1+\Phi'(\qt))$.
    Since $\bn_m=\frac{1}{K} \sum_{t\in\C{T}_m} \nabla_t$, using Cauchy-Schwartz inequality, we have $\Vert \bn_m \Vert ^2 = \frac{1}{K^2} \Vert \sum_{t\in\C{T}_m} \nabla_t \Vert ^2 \leq \frac{1}{K} \sum_{t\in\C{T}_m}\normdt^2 $. Thus,
    \begin{align}
        \label{eq:sqrt-sum-gm}
        \sqrt{\sum_{m=1}^{T/K} \Vert \bn_m \Vert ^2} 
        &\leq \sqrt{\frac{1}{K} \sum_{m=1}^{T/K} \sum_{t\in\C{T}_m}\normdt^2} \nonumber\\
        &= \sqrt{ \frac{1}{K} \sum_{t=1}^{T} \normdt^2 } \nonumber \\
        &\overset{(a)}\leq \frac{\gamma M_1}{\sqrt{K}}\sqrt{ \sum_{t=1}^{T} (1+\Phi'(\qt))^2 } \nonumber \\
        &\overset{(b)}\leq \frac{\gamma M_1}{\sqrt{K}}\sqrt{ \sum_{t=1}^{T} 2(1+\Phi'(\qt)^2) } \nonumber \\
        &\overset{(c)}\leq \frac{\gamma M_1 \sqrt{2}}{\sqrt{K}} (\sqrt{T}+\sqrt{\sum_{t=1}^{T}\Phi'(\qt)^2})\nonumber \\
        &\overset{(d)}\leq \frac{\gamma M_1 \sqrt{2T}}{\sqrt{K}} (1 + \Phi'(Q_T))
    \end{align}
    where (a) follows from Equation~\eqref{eq:Gt} and the fact that $G_t=\gamma M_1(1+\Phi'(\qt))$, (b) follows from Cauchy-Schwartz inequality, (c) follows from Jensen's inequality, and (d) is because $\Phi'(\cdot)$ is non-decreasing as $\Phi(\cdot)$ is convex and $Q_t$ is non-decreasing, thus $\Phi'(Q_t) \leq \Phi'(Q_T), \forall t\in[T]$.
    By plugging Equation~\eqref{eq:Gt} and Equation~\eqref{eq:sqrt-sum-gm} into Lemma~\ref{lem:surrogate-regret} and rearranging terms, we have Equation~\eqref{eq:sur2real:2} in Lemma~\ref{lem:sur2real}.
\end{proof}

Having necessary lemmas at hand, we now move towards parameter tuning.
If we set $\Phi(Q_T)=e^{\lambda Q_T}-1$ with some real number $\lambda$, we have $\Phi'(Q_T)=\lambda e^{\lambda Q_T}$. Letting $S \triangleq \delta T + \frac{3}{\sqrt{2}}\sqrt{TK}$, we have 
\begin{align}
\label{eq:regretTft}
    &\mathtt{R}(\tf)  \overset{(a)}{\leq} \mathtt{R} (\hf)-\Phi (Q_T) \nonumber \\
    &\quad\overset{(b)}{\leq}
    \gamma M_1 D (V+\lambda e^{\lambda Q_T})(\delta T + \frac{3}{\sqrt{2}}\sqrt{TK}) + \frac{KD}{2}\sqrt{\epsilon}
    -e^{\lambda Q_T}+1\nonumber\\
    &\quad\overset{(c)} = \gamma M_1 D S 
    + e^{\lambda Q_T} \left( \lambda \gamma M_1 D S-1 \right)+1 + \frac{KD}{2}\sqrt{\epsilon}
\end{align}
where (a) uses Equation~\eqref{eq:sur2real:1}, (b) uses Equation~\eqref{eq:sur2real:2}, and (c) rearranges terms and adopts $S$.

We make a few observation of Equation~\eqref{eq:regretTft}.
Note that if we take $\gamma=(M_1D)^{-1}$, we can offset $\gamma M_1 D$ term entirely.  
If we choose $\lambda=\frac{1}{2S}$, then $\lambda \gamma M_1 D S -1 = -\frac{1}{2}<0$. 
Thus, based on Equation~\eqref{eq:regretTft}, we have
\begin{align}
    &\mathtt{R}(\tf) \leq S+1= \delta T + \frac{3}{\sqrt{2}}\sqrt{TK}+1 + \frac{KD}{2}\sqrt{\epsilon}.
\end{align}
Since $\tft=\gamma f_t$, we have $\mathtt{R}(\tf) = \gamma\mathtt{Regret}(T)$. 
Let $\beta\in[0,\frac{1}{2}]$ be some real number. Thus, if we choose $\delta=\Theta(T^{-\beta})$ and $K=\Theta(T^{1-2\beta})$,
\begin{align*}
    \mathtt{Regret}(T) 
    &= \gamma^{-1} \mathtt{R}(\tf) \\
    &= D M_1 \delta T + \frac{3}{\sqrt{2}}DM_1\sqrt{TK}+DM_1 + \frac{K D^2 M_1}{2}\sqrt{\epsilon}\\
    &= \C{O}(DM_1 T^{1-\beta}).
\end{align*}

We now take a closer look at the  Cumulative Constraint Violation (CCV). With $\tft=\gamma f_t$ and $f_t$ being $M_1$-Lipschitz, we have $\tft$ is $\gamma M_1$-Lipscitz continuous, and thus $\tft(\xt)-\tft(\xs) \geq -\gamma M_1 (\xt-\xs) \geq -1$. Taking sum over $t\in[T]$, we have
\begin{align}
\label{eq:lowbon4regret}
    \mathtt{R}(\tf)=\sumT(\tft(\xt)-\tft(\xs)) \geq -T
\end{align}
Replacing $\mathtt{R}(\tf)$ in Equation~\eqref{eq:regretTft} with the lower bound in Equation~\eqref{eq:lowbon4regret} and rearranging terms, we have
\begin{align*}
     e^{\lambda Q_T} \le \frac{S + 1 + T + \frac{KD}{2}\sqrt{\epsilon}}{1 - \lambda S} 
\end{align*}
Since $\lambda=\frac{1}{2S}$, we have
\begin{align*}
Q_T &\leq 2S \log\left( 2(S + 1 + T + \frac{KD}{2}\sqrt{\epsilon})\right) 
\end{align*}

Considering that $Q_T
= \sum_{t=1}^T \widetilde g_t(x_t)
= \gamma \sum_{t=1}^T (g_t(x_t))_+
= \gamma V(T)$, with $S = \delta T + \frac{3}{\sqrt{2}}\sqrt{TK}$ and $\gamma=(DM_1)^{-1}$, if we choose $\delta=\Theta(T^{-\beta})$ and $K=\Theta(T^{1-2\beta})$ for some $\beta\in(0,\frac{1}{2}]$, we have
\begin{align*}
    \mathtt{CCV}(T) \le V(T) &= \gamma^{-1} Q_T 
    \leq 2DM_1S * \C{O}(\log T) \\
    &\leq 2 D M_1 (\delta T + \frac{3}{\sqrt{2}}\sqrt{TK}) * \C{O}(\log T) \\
    &\leq \C{O}(DM_1T^{1-\beta}\log T)
\end{align*}

Since calls to the infeasible projection and separation oracles are done in $\C{A}$, we can extend results from Corollary~\ref{cor:con} and derive that the oracle calls are bounded as $\tilde{\C{O}}(T^{2\beta})$.


\section{Analysis of BAGEL for Strongly Convex Costs}
\label{apx:constrained-strongly}

We note that the regret bounds can improve when the cost functions are strongly convex. In this section, we describe how the proposed algorithm, $\mathtt{BAGEL}$, can be implemented when the cost functions are strongly convex, and analyze the corresponding $\mathtt{Regret}$, $\mathtt{CCV}$, and total number of calls to the separation oracle guarantees. 

Let the cost functions $f_t$ be $\theta$-strongly convex while the constraint functions are considered to be convex, not necessarily strongly convex. Let them all be $M_1$-Lipschitz continuous. 
In this case, we still apply $\mathtt{BAGEL}$ (Algorithm~\ref{alg:constrained}) with different parameters. In line~10, we add a regularization parameter $V$ to control the influence of costs on the surrogate function, i.e., we let $\hat f_t \gets V\,\tilde f_t + \Phi'(Q_t)\,\tilde g_t$. Besides, we choose \(\gamma = 1\), \(V = \frac{8M_1^2 K \log(Te/K)}{\theta}\), and \(\Phi(\cdot) = (\cdot)^2\). Further, the Base Algorithm instance $\C{A}$ uses step size \(\eta_m = \frac{1}{m\theta}\). For clarity, we include the updated BAGEL algorithm for strongly-convex costs in Algorithm~\ref{alg:strongly-convex}.

\begin{algorithm}[H] 
\caption{\textsc{BAGEL} for Strongly-convex costs}
\label{alg:strongly-convex}
\begin{algorithmic}[1]
  \Require Action set $\K$, time horizon $T$, $\BF{c} \in \op{relint}(\K)$, $r = r_{\K, \BF{c}}$, shrinking parameter $\delta \in [0,1)$, block size $K$, action set diameter $D$, an instance of the Base Algorithm (Algorithm~\ref{alg:general}) $\C{A}$, processing parameter $\gamma$, regularization parameter $V$, Lyapunov function $\Phi(\cdot)$, an instance of IP-SO (Algorithm~\ref{alg:ipso}) $\C{P}$, strongly-convex parameter $\theta$. 
  \State \textbf{Initialization:} Pick $\mathbf{x}_1\in\mathcal{K}$; set $Q_0\gets 0$; pass $\mathcal{K},T,K,\mathbf{c},r,\delta,D,\mathcal{P}$, and $(\epsilon\text{ or }\theta)$ to $\mathcal{A}$
  \For{$m \in [T/K]$}
    \For{$t\in\mathcal{T}_m$}
      \State Play $\mathbf{x}_m$
      \State Adversary reveals $f_t$ and $g_t$
      \State Observe $f_t(\mathbf{x}_m)$ and $(g_t(\mathbf{x}_m))^{+}$
      \State $\tilde f_t \gets \gamma\, f_t$\;; \quad $\tilde g_t \gets \gamma\, (g_t)^{+}$
      \State $Q_t \gets Q_{t-1} + (g_t(\mathbf{x}_m))^{+}$
      \State $\hat f_t \gets V\,\tilde f_t + \Phi'(Q_t)\,\tilde g_t$\;; \quad $\nabla_t \gets \nabla \hat f_t(\mathbf{x}_m)$
      \State Pass $\nabla_t$ to $\mathcal{A}$
    \EndFor
    \State Receive $\mathbf{x}_{m+1}$ from $\mathcal{A}$
  \EndFor
\end{algorithmic}
\end{algorithm}

Having studied the regret for OCO for strongly convex functions in Section~\ref{sec:oco}, we now use the result to study the regret and CCV for the adversarial COCO problem for strongly convex costs, with similar approach to the proofs in Section~\ref{sec:pacogd}. The results are given in Theorem~\ref{thm:constrained-strongly}, and the detailed proof can be found in Appendix~\ref{apx:constrained-strongly}.

\begin{theorem}
\label{thm:constrained-strongly}
    Assume cost functions $f_t$ are $\theta$-strongly convex and constraint functions $g_t$ are convex, and they are all $M_1$-Lipschitz continuous.
    With $\beta\in (0,1]$, if we choose $\gamma=1$, $\delta= \Theta( T^{-\beta} \log(T) )$, $K= \Theta(T^{1-\beta})$, $V=\frac{8 M_1^2 K \log(Te/K)}{\theta}$, and $\Phi(\cdot)=(\cdot)^2$,
    Algorithm~\ref{alg:constrained} ensures that
    \begin{align*}
        \mathtt{Regret}(T) &= \C{O}(T^{1-\beta}\log(T)), \\
        \mathtt{CCV}(T) &= \C{O}(T^{1-\frac{\beta}{2}}\sqrt{\log(T)}),
    \end{align*}    
    with at most $\C{O}(T^{2\beta}(\log(T))^{-2})$ calls to the separation oracle.
\end{theorem}
\begin{remark}
    If we take $\beta=1$, we obtain optimal results $\log(T)$ for $\mathtt{Regret}$ and $\sqrt{T\log(T)}$ for $\mathtt{CCV}$ if permitted $\C{O}(T^2)$ calls to the separation oracle, which matches the optimal result for projection-based algorithm in \citet{sinha2024optimal}. If only $\C{O}(T)$ calls permitted, we take $\beta=\frac{1}{2}$, and still have $\C{O}(\sqrt{T})$ regret and $\C{O}(T^{\frac{3}{4}})$ CCV, which shows significant improvement compared to $\C{O}(T^\frac{3}{4})$ regret and $\C{O}(T^{\frac{7}{8}})$ CCV in \citet{garber2024projection}. 
\end{remark}
\begin{proof}
    We observe that Equation~\eqref{eq:sur2real:1} still holds when $f_t$ are strongly convex. Similar to proof of Theorem~\ref{thm:main}, we first derive bound on $\mathtt{R}(\hf)$, and move to $\mathtt{R}(\tf)$, before obtaining results on $\mathtt{Regret}(T)$. Then we bound regret from below and choose parameters appropriately to obtain results for $\mathtt{CCV}(T)$.

    If $\gamma=1$, we have $\tft=f_t$, $\tgt=(g_t)^+$. Thus, $\hft(\xt)=V\tft(\xt)+\Phi'(\qt)\tgt=Vf_t+\Phi'(\qt)(g_t)^+$, where $V$ is a positive real number, and $\Phi(\cdot)$ is a Lyapunov function as described in Section~\ref{sec:pacogd}. 
    Since $g_t$ are convex, $(g_t)^+$ are also convex. 
    Since $\Phi(\cdot)$ is non-decreasing, $\Phi'(\qt)$ is non-negative.
    As $f_t$ are $\theta$-strongly convex, we can apply Lemma~\ref{lem:scalar-stronvex} and derive that $\hft(\xt)$ are $V\theta$-strongly convex.
    Since both $f$ and $g$ are assumed to be $M_1$-Lipschitz continuous, Equation~\eqref{eq:Gt} still holds: $\normdt \leq \gamma M_1(V+\Phi'(\qt))$. Thus, surrogate functions $\hft$ are $G_t$-Lipschitz continuous, where $G_t = \gamma M_1(V+\Phi'(\qt))$.
    Similar to how we extend Theorem~\ref{thm:unconstrained-regret} to get Lemma~\ref{lem:surrogate-regret}, we can extend Theorem~\ref{thm:unconstrained-strongly} and get:
    \begin{align*}
        \mathtt{R}(\hf) & \overset{(a)}\leq \frac{1}{2V\theta}\sum_{m=1}^{T/K}\sum_{t\in\C{T}_m} \frac{\normdt^2}{m} + \frac{\delta D}{r} \sumT G_t\\
        & \overset{(b)} \leq \frac{1}{2V\theta} \sum_{m=1}^{T/K}\sum_{t\in\C{T}_m} \frac{M_1^2(V+\Phi'(\qt))^2}{m} \\
        & \qquad+ \frac{\delta D}{r} \sumT M_1(V+\Phi'(\qt))  \\
        & \overset{(c)} \leq \frac{VM_1^2K}{\theta}(1+\log(\frac{T}{K})) \\
        & \qquad+\frac{M_1^2}{V\theta } \sum_{m=1}^{T/K}\sum_{t\in\C{T}_m} \frac{(\Phi'(\qt))^2}{m} \\
        & \qquad+ \frac{\delta D}{r} M_1 \sumT (V+\Phi'(\qt))
    \end{align*}
    where (a) extends Theorem~\ref{thm:unconstrained-strongly} and uses the fact that $\hft(\xt)$ are $V\theta$-strongly convex, (b) follows from Equation~\eqref{eq:Gt} and the fact that $G_t = \gamma M_1(V+\Phi'(\qt))$, and (c) uses the $1+\log(n)$ bound on sum of harmonic series with $n$ terms.
    Since Equation~\eqref{eq:sur2real:1} still holds, taking $\Phi(\cdot)=(\cdot)^2$, we have
    \begin{align*}
        &Q_T^2+V\mathtt{Regret}(T) \leq \mathtt{R}(\hf) 
        \leq \frac{VM_1^2K}{\theta}(1+\log(\frac{T}{K})) \\
        & \qquad + \frac{4M_1^2}{\theta V } \sum_{m=1}^{T/K}\sum_{t\in\C{T}_m} \frac{\qt^2}{m} 
        + \frac{\delta D}{r} M_1 \sumT (V+2\qt).
    \end{align*}
    Since $Q_t$ is non-decreasing, we have $Q_T \geq Q_t \geq 0, \forall t\in[T]$. With the sum of harmonic series $\sumT \frac{1}{t} = \log(Te)$, and $V$ being a positive real number, we have
    \begin{align}
    \label{eq:stronvex-reg}
        &\mathtt{Regret}(T) \leq \frac{M_1^2K}{\theta}(1+\log(\frac{T}{K})) \nonumber \\
        & \quad +\frac{4M_1^2KQ_T^2 \log(Te/K)}{\theta V^2} 
        + \frac{\delta D}{r} M_1 T (1+\frac{2Q_T}{V}) - \frac{Q_T^2}{V} \nonumber\\
        & \leq \frac{M_1^2K}{\theta}(1+\log(\frac{T}{K})) 
         +\frac{4M_1^2KQ_T^2 \log(Te/K)}{\theta V^2} \nonumber \\
        & \quad + \frac{\delta D}{r} M_1 T (1+\frac{2Q_T}{V}). 
    \end{align}
    If we choose $\delta=\Theta(T^{-\beta} \log(T))$ and $K=\Theta(T^{1-\beta})$, where $\beta \in (0,1]$ is a trade-off parameter,we have 
    \begin{align*}
       \mathtt{Regret}(T) = \C{O}(T^{1-\beta}\log(T)),
    \end{align*}
    and hence we prove the result for $\mathtt{Regret}$ in Theorem~\ref{thm:constrained-strongly}.
    
    Next we take a look at $\mathtt{CCV}(T)$. Given that the cost functions are $M_1$-Lipschitz continuous, we have 
    \begin{align*}
        \mathtt{Regret}(T)&=\sumT(f_t(\xt)-f_t(\xs)) \\
        &\geq \sumT-M_1\Vert \x_t - \xs \Vert \\
        &\geq -M_1DT.
    \end{align*}
    Replacing $\mathtt{Regret}(T)$ in Equation~\ref{eq:stronvex-reg} with the above lower bound, and choose $V=\frac{8M_1^2K\log(Te/K)}{\theta}$, we have,
    \begin{align*}
        \frac{1}{2}Q_T^2-\frac{2\delta D}{r}M_1 T Q_T 
        &\leq V(\frac{\delta D}{r} M_1 T + D M_1 T \\
        &\quad+ \frac{M_1K}{\theta}(1+\log(\frac{T}{K}))).
    \end{align*}
    Let $B_T\triangleq V(\frac{\delta D}{r} M_1 T + D M_1 T + \frac{M_1K}{\theta}(1+\log(\frac{T}{K})))$, then we have 
    \begin{align*}
        Q_T^2- \frac{4\delta D}{r} M_1 T Q_T - 2B_T \leq 0.
    \end{align*}
    Note that the left hand side is a quadratic function of $Q_T$.  Solving the inequality, we have
    \begin{align*}
        Q_T \leq \frac{\frac{4\delta D}{r} M_1 T + \sqrt{(\frac{4\delta D}{r} M_1 T)^2 + 8B_T}}{2}.
    \end{align*}
    Since $B_T = \C{O}(T^{2-\beta}\log(T))$, we have $Q_T = \C{O}(T^{1-\frac{\beta}{2}}\sqrt{\log(T)})$, and $\mathtt{CCV}(T)= \sumT (g_t)^+ = Q_T = \C{O}(T^{1-\frac{\beta}{2}}\sqrt{\log(T)})$ for $\theta$-strongly convex cost. 

    Since calls to the infeasible projection and separation oracles are done in $\C{A}$, we can directly extend results from Corollary~\ref{cor:strcon} and bound oracle calls as $\C{O}(T^{2\beta}(\log(T))^{-2}))$.
\end{proof}

\end{document}